\documentclass{article}

\usepackage{arxiv}

\usepackage[utf8]{inputenc} 
\usepackage[T1]{fontenc}    
\usepackage{hyperref}       
\usepackage{url}            
\usepackage{booktabs}       
\usepackage{amsfonts}       
\usepackage{amsmath}
\numberwithin{table}{section}

\DeclareMathOperator*{\argmax}{argmax} 

\usepackage{nicefrac}       
\usepackage{microtype}      
\usepackage{lipsum}

\usepackage{caption} 
\usepackage{amsmath}
\usepackage{amssymb}

\numberwithin{equation}{section}
\numberwithin{figure}{section}

\usepackage{url} 
\usepackage[table]{xcolor}
\definecolor{dgreen}{rgb}{0.0, 0.5, 0.0}
\definecolor{dred}{rgb}{0.8, 0.0, 0.0}
\usepackage{float}

\usepackage[linesnumbered,algoruled,boxed,lined]{algorithm2e}
\usepackage{array,multirow,graphicx}

\usepackage{placeins}

\let\Oldsection\section
\renewcommand{\section}{\FloatBarrier\Oldsection}

\usepackage{natbib}

\title{A Reinforcement Learning Approach to the Orienteering Problem with Time Windows}

\author{
  Ricardo Gama\\
  Research Centre in Digital Services (CISeD)\\
Polytechnic Institute of Viseu, Portugal\\
  \texttt{rgama@estgl.ipv.pt} \\
   \And
 Hugo L. Fernandes \\
  Rockets of Awesome\\
  New York City, USA\\
  \texttt{hugoguh@gmail.com} \\
}
\date{\today}

\begin{document}
\maketitle

\begin{abstract}
The Orienteering Problem with Time Windows (OPTW) is a combinatorial optimization problem where the goal is to maximize the total score collected from different visited locations. The application of neural network models to combinatorial optimization has recently shown promising results in dealing with similar problems, like the Travelling Salesman Problem. A neural network allows learning solutions using reinforcement learning or supervised learning, depending on the available data. After the learning stage, it can be generalized and quickly fine-tuned to further improve performance and personalization. The advantages are evident since, for real-world applications, solution quality, personalization, and execution times are all important factors that should be taken into account. 

This study explores the use of Pointer Network models trained using reinforcement learning to solve the OPTW problem. We propose a modified architecture that leverages Pointer Networks to better address problems related with dynamic time-dependent constraints. Among its various applications, the OPTW can be used to model the Tourist Trip Design Problem (TTDP). We train the Pointer Network with the TTDP problem in mind, by sampling variables that can change across tourists visiting a particular instance-region: starting position, starting time, available time, and the scores given to each point of interest. Once a model-region is trained, it can infer a solution for a particular tourist using beam search. We based the assessment of our approach on several existing benchmark OPTW instances. We show that it generalizes across different tourists that visit each region and that it generally outperforms the most commonly used heuristic, while computing the solution in realistic times.
\end{abstract}

\section{Introduction}

 The Orienteering Problem and its variants are a family of combinatorial optimization problems with numerous applications, from logistics to telecommunications (\cite{OPbook}). Among them, the Orienteering Problem with Time Windows (OPTW) can be seen as a model for the Tourist Trip Design Problem (TTDP) (\cite{Gavalas2014}). When they plan a trip to a given destination, tourists intend to visit most of their favourite points of interest within a limited time schedule. Usually, the large number of points of interest, located in different places and with different operation hours, makes it arduous and time-consuming to design relevant tourist routes. In the OPTW problem, one has to find the route that maximizes the sum of scores of the visited points of interest considering local and global time constraints. 

The OPTW is an NP-hard problem and several heuristics have been proposed to obtain high-quality solutions. The Iterate Local Search (ILS) (\cite{Vansteenwegen2009})heuristic is one of the most well-known and provides fast and good quality solutions, which allows its incorporation in real-time applications. \cite{Gunawan2015} extended the ILS approach making some point-wise adaptations and including other local search operations, such as swap, 2-opt and replace, which improved the score on standard benchmark instances at the expense of computation time. More recently, large neighbourhood search (\cite{Schmid2017}) and evolution strategy approaches (\cite{KARABULUT2020106109}) were found to improve best-known results on several benchmark instances. Several other local search-based heuristics have been proposed to tackle OPTW, like variable neighbourhood search methods, ant colony optimization or simulated annealing heuristics (see e.g. \cite{Vansteenwegen2011, Gunawan2016, OPbook}). For practical applications, the heuristic that will be chosen depends on the trade-off between the aimed quality of the solution and execution time constraints. Several heuristics have been proposed for the OPTW yet, compared to machine learning models, a typical heuristic has smaller potential for generalization and personalization.

 Recently, there have been some encouraging results showing the possibility of solving combinatorial optimization problems using neural networks  (\cite{Bengio2018}, \cite{Dai2017}). In the work of \cite{Vinyals2015} the authors introduced a neural network model called Pointer Network (PNs) and applied it to three classical combinatorial problems: planar convex hulls problem, Delaunay triangulation, and the planar Travelling Salesman Problem (TSP), with very promising results. One important aspect of their approach is that their model is trained using supervised learning on generated training data. This may be advisable when ground-truth data is promptly accessible but may be a drawback for some combinatorial optimization problems for which data collection is not possible. The lack of real data for model training imposes the existence of efficient heuristics to generate approximated ground-truth data and leads to a model performance strongly influenced by the performance of the selected heuristics. Furthermore, in most problems, fast solutions can only be obtained for smaller problems or at the expense of the quality of the solution. This limits the applicability of the proposed approaches to smaller size problems and leaves out important practical applications. 

To overcome these limitations, \cite{Bello2016} used reinforcement learning to train Pointer Networks for the TSP and the Knapsack Problem. This opened the way to the possibility of tackling a broader set of combinatorial optimization problems. Following that work, other Pointer Network architectural modifications and applications have appeared, e.g., Vehicle Routing Problems ( \cite{Deudon2018}, \cite{Nazari2018}, \cite{Kool2019a}, \cite{Falkner2020} and \cite{Lin2020}), Knapsack Problem (\cite{Gu2018}) or the Max-Cut Problem (\cite{Gu2018b}).

An optimization problem to which the use of these machine learning techniques can be particularly beneficial is the TTDP. In a realistic setting, we can expect the variability across instances of the same region to come mostly from the choices made by tourists, and not so much from region specific characteristics like the location of its points of interest or their opening and closing times. This means that, in practice, one needs to repeatedly solve a nearly same instance problem where only a limited set of specific features/parameters may vary, usually in an expected and controlled way. Thus, when access to real data is not possible, the generation of realistic instances for training can be achieved through simulation. This should allow the training with reinforcement learning of a model that can generalize across that variability.

 In this paper, we explore the application of Pointer Networks to the OPTW/TTDP problem. We propose a model that takes into account the recurrent nature of the construction process of the sequential solution, using Pointer Networks to better address problems with dynamic time-dependent constraints. This is a recursive attentive model that relies on a Transformer block (\cite{Vaswani2017, Deudon2018, Kool2019a}) with dynamical graph self-attention (\cite{Velickovic2018}). We use reinforcement learning to train the model for a particular instance-region (a fixed set of points of interests with fixed coordinates, opening and closing times and duration of visit). During learning, we generate new instances (new instant-tourists) for that instance-region that vary depending on the score provided by tourists to each point of interest, starting time and position, and the time available for the tour. We generate these instances in a way that mimics the variability expected across tourists so that we might obtain a model capable of generalizing across tourists who visit that region/city/neighbourhood.

\section{Orienteering Problem with Time Windows}
\label{sec:optw}

In an OPTW instance, $\phi$, a set of $n$ nodes, $\{ \mathrm{v}_i \}_{i=1}^n$ with their corresponding coordinates  $x_i \in \mathbb{R}^2$ are given. Every node $\mathrm{v}_i$ has a positive score or reward, $r_i$, a visiting time window $[o_i , c_i]$ with opening time (the earliest a visit can start) and closing time (the time at which a visit has to stop), and duration of visit $d_i$. Without loss of generality, we can assume that the starting location is $\mathrm{v}_1$, and $\mathrm{v}_n$ is the ending location for every solution path (also called route). The objective is to find a solution route $ S = (\pi_1, \ldots, \pi_m)$ with the maximum possible sum of scores, without repeating visits, starting the route at or after a given time $T_{start}$ and ending it before time $T_{end}$ (see \cite{OPbook}).

We build a solution in a sequential fashion. For each step of the process, the existing time budget and time windows constraints have to be taken into account. We start the route in $v_1$, initializing the current time $t^0$ with the particular instance starting time $T_{start}$. After the first initialization step, the path construction process follows iteratively, setting which node,  $\mathrm{v}_j$, will be visited next assuming that the following constrains are satisfied:
\begin{equation}
\label{eq:fcond1}
a_j+wait_j  \leq c_j;
\end{equation}\begin{equation}
\label{eq:fcond2}
a_j+wait_j + d_j + \Delta_{jn} \leq T_{end},
\end{equation}
where $a_j$ is the time of arrival at node $\mathrm{v}_j$, $\text{wait}_j = \max (0, o_j - a_j)$ is the time one has to wait before the visit can start, and $\Delta_{jn}$ the time  it takes to go from node $\mathrm{v}_j$ to node $\mathrm{v}_n$. That is, one must arrive at $\mathrm{v}_j$ before closing time and, at arrival, there needs to be enough time left to visit the $\mathrm{v}_j$ and travel to the last location $\mathrm{v}_n$.  After the visit to node $\mathrm{v}_j$, the current time is updated $t^l = a_{j}+wait_{j} + d_{j}$ and the construction loop proceeds until $\mathrm{v}_n$ is reached.

At each iteration step $l$, we define $\mathcal{A}^l(\mathrm{v}^l_{\ast}, t^l)$ as the set of nodes that are admissible to be visited next. This set is a function of the sub-sequence built until step $l$, and of the current node $\mathrm{v}^l_{\ast}$ and current time $t^l$.

\subsubsection*{The Tourist Trip Design Problem Perspective}
An OPTW instance can be interpreted as a TTDP instance, i.e. a particular tourist visiting a particular region. From this perspective, a route can be considered a tourist tour, and the nodes will be points of interest that the tourist might want to visit. Some of the parameters are instance-region specific and tourist invariant, while others are not. Here, we assume that a route’s starting location $\mathrm{v}_1$, its starting time $T_{start}$ and ending time  $T_{end}$,  as well as the scores obtained by the different points of interest are tourist dependent, while the coordinates of the points of interest, their opening and closing times, and the duration of the visit are tourist invariant. We use 3 different sets of benchmark instances (\cite{Vansteenwegen2011}, \cite{Gavalas2019}, see Section \ref{subsec:benchmark_instances}).Each benchmark instance can be interpreted as a particular tourist visiting a particular instance-region. For each of these instances we can generate new tourists with some variability across the tourist-dependent parameters (see Section \ref{subsec:generated_instances} in  Methods). We use these generated tourists or tourist-instance-regions to train, monitor, evaluate and validate our model.

\section{Pointer Network Model}
\label{sec:review}

Our model is a graph recursive attentive model based on the Pointer Network model architecture (\cite{Vinyals2015}). We conducted a detailed description of the model and of the Pointer Network architecture to explain the particularities of the model, the differences in relation to the other approaches used to solve combinatorial optimization problems using Pointer Network models, and the motivation behind those choices.

\begin{figure}[ht]
\includegraphics[width=0.65\textwidth]{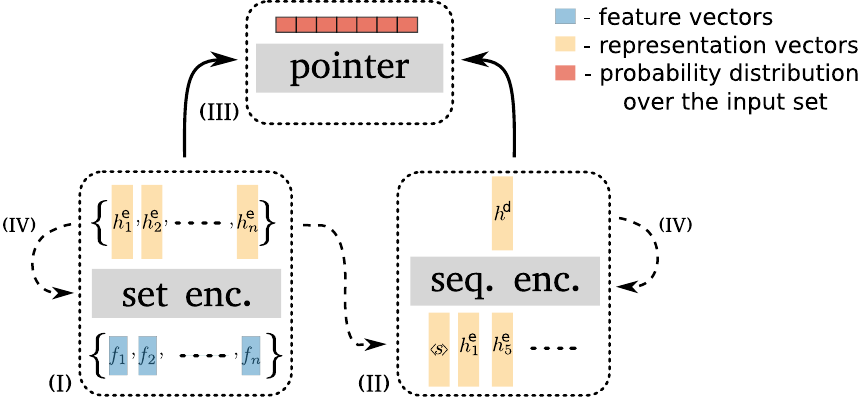}
\centering
\caption{The proposed Pointer Network architecture and information flow during an iteration step.  (I) - Set encoder block;  (II) - Sequence encoder block;  (III) - Pointing mechanism; (IV) - Representation vectors computed in the previous iteration step are used recursively in their corresponding functional block.}
\label{Pointnet_fig}
\end{figure}

For a large set of combinatorial optimization problems, e.g. the TSP and the OPTW, a solution can be constructed sequentially, with elements chosen iteratively one by one. The Pointer Network model architecture (\cite{Vinyals2015}) was designed to address this kind of problem. Its architecture consists of tree main blocks (see Figure \ref{Pointnet_fig}): the \textit{set encoder} block that processes the input set; the \textit{sequence encoder} block that handles the sequence as it is built; and a third block that consists of a \textit{pointing mechanism}. Typically, this sequential process is carried out in two distinct phases: an encoding phase and a pointing phase. During the encoding phase, the set encoder block creates a higher-dimensional feature representation of the given input set. Afterwards, the sequence encoding block encodes the solution/sub-sequence constructed so far. In the second phase, both representations are fed into the pointing mechanism to generate a probability distribution over the input set. This probability distribution "points" to which element is more likely to be the best choice as the next sequence element.

Both the encoding (set and sequence) and the pointing mechanisms can be design in different ways (e.g.\cite{Vinyals2015, Bello2016, Nazari2018, Kool2019a}).
Here we use the same sequence encoding and pointing mechanism architecture as \cite{Vinyals2015}. The main and substantial differences regarding this and other existing Pointer Network variations lie in the set encoding block. There are three main aspects to highlight when comparing previous Pointer Network studies. Before going into more detail, we will briefly outline them.

\textbf{New set representation at every iteration.} 

The first essential aspect is that the set encoder runs in parallel with the sequence encoder and provides an updated vector representation of each node in every iteration step. When building a solution, at each iteration, we recompute the representation of each node (the output of the set encoding block), feeding both dynamic and static features to the set encoding block (\cite{Dai2017}). 

It is important to note that this is different from previous proposed Pointer Network models, including from those that also use both static and dynamic features (\cite{Nazari2018, Kool2019a}). In those works, the static features are only encoded once: in the first step of the solution building process. The dynamic features are used by the sequence-encoder block and pointing mechanism which rely on context-based attention.

In fact, the inclusion of this adjustment in the model architecture allows us to: use recursion in the set encoding block (Transformer with recursion, see below); mask the set of nodes and lookahead connections that are not admissible (Masked self-attention using induced graph structure, see below).

\textbf{Transformer with recursion.} A second aspect is that we use self-attention with recursion (see Section \ref{sec:transformer_recursion} for details, and “no recursion” in Supplementary materials \ref{app:notweaks} for model comparison) to try to make the best of the recurrent nature of the solution building process. Intuitively, this change somewhat brings part of the sequence encoding effort to the set encoding block. 

\textbf{Masked self-attention using induced graph structure.} A third important aspect is that at each iteration step, and in order to consider the time constraints of the problem, we dynamically compute the set of admissible nodes and a graph structure that can be induced over it (see Section \ref{sec:graph} for detains, and “complete graph” in Supplementary materials \ref{app:notweaks} for model comparison). We use this graph-structure to apply masked self-attention during set encoding.

\subsection{Set Encoding}
\label{sec:set_encoding_block}
Formally, given an input set of nodes $\mathrm{V} = \lbrace \mathrm{v}_1,  \mathrm{v}_2,\ldots,  \mathrm{v}_n \rbrace$, and their corresponding feature vectors $f = \lbrace f_1,  f_2,\ldots,  f_n \rbrace$, where $f_i \in \mathbb{R}^{d_f}$, the first step is to project these vectors into a higher-dimension space through a learnable transformation.

We separate $f_i$ into static $f^{st}_i$ and dynamic $f^{dy}_i$ groups (\cite{Nazari2018} ).  As the model is building a particular solution, the dynamic features (e.g. time until $\mathrm{v}_i$ opens) may change at each iteration, while the static remain the same throughout that process (e.g. Euclidean coordinates of $\mathrm{v}_i$). We map each $f_i$ into $e_i \in \mathbb{R}^{d_{e}}$ where $e_i = [e^{st}, e^{dy}]$, $e^{st}_i = \tanh{(W^{st} f^{st}_i + b^{st})}$and $e^{dy}_i = \tanh{(W^{dy} f^{dy}_i + b^{dy})}$, i.e. in a first layer we process the static and dynamic features separately before concatenating them.

These embedded node vectors $e =\lbrace e_1,  e_2,\ldots,  e_n \rbrace$ are fed into a Transformer (\cite{Vaswani2017}) with a tweak. This tweak is the introduction of recursion in computing the \textit{key} of the self-attention sub-block (see details below). The final output of the set encoder and of the transformer is a set $\lbrace h^e_1,  h^e_2,\ldots,  h^e_n  \rbrace$ representation of $\mathrm{V}$, with $h^e_i \in \mathbb{R}^{d_{e}}$. We chose the Transformer architecture both because it is suitable for permutation invariant input sets (the set of points of interest) and because of the good results reported in similar combinatorial optimization problems (\cite{Deudon2018, Kool2019a}).

\subsubsection{Vanilla Transformer}
\label{sec:vanilla}

In this section we describe the original transformer architecture before explaining how we introduce recursion. The Transformer architecture consists of a set of stacked layers (here we use 2 layers). Each layer is composed of 2 sublayers or functional components: a multi-head self-attention sublayer followed by a feedforward sublayer.

Given a list of input vectors $\lbrace e_1,  e_2,\cdots,  e_n  \rbrace$, the self-attention vectors are determined (note that this is the description of the vanilla Transformer, in our model the keys $k_i$ are computed recursively, see Section \ref{sec:transformer_recursion}) by first computing two linear transformations for each $e_i$: 
\begin{equation}
\label{eq:self_attention_key}
k_i = W_k e_i
\end{equation}
\begin{equation}
\label{eq:self_attention_value}
v_i = W_v e_i,
\end{equation}
followed by the determination of the similarity score between each $e_i$ and each $k_j$, using scaled dot-product attention:
\begin{equation}
\label{eq:simscore}
u_{ij} = \dfrac{e_i^T W_q^T  k_j}{\sqrt{d_k}}, \qquad \forall i,j \in \{ 1,\ldots, n \},
\end{equation}
where $d_k$ is a scaling constant,  $W_k$,   $W_v$ and $W_q  \in \mathbb{R}^{d_{h^a} \times d_e}$. The variables $k_i$ and $v_i$ in Equations \ref{eq:self_attention_key} and \ref{eq:self_attention_value} are usually called \textit{key} and \textit{value}, respectively.
The output vector $h_i^a$ is then computed by taking the average of the vector of values $v$, using a vector $a_i$, the softmax normalization of the similarity scores vector $u_i$:
\begin{equation}
h^a_i= \sum_{j} a_{ij} v_j , \quad \text{with} \quad a_{ij} = \dfrac{\exp (u_{ij})}{\sum_{j^{\prime}} \exp (u_{ij{\prime}})}.
\end{equation}
For multi-head attention, the attention mechanics is independently applied, i.e without weight sharing, multiple times in parallel to the same input. Here we used 8 heads. The vectors resulting from each attention head are concatenated and once again projected into $\mathbb{R}^{d_e}$ through a linear transformation. This results in the attention representation vectors $\lbrace h^a_1,  h^a_2,\cdots,  h^a_n  \rbrace$. 

The second component of the Transformer layer is then a feedforward fully connected layer:

\begin{equation}
\label{eq:ff}
h_i = W_2^{ff}  \text{Relu}(W_1^{ff} h^a_i + b_1^{ff})  + b_2^{ff}
\end{equation}

where $\text{Relu}(x) = \max(0,x)$, $W_1^{ff}\in \mathbb{R}^{d_{ff} \times d_{e}}$,  $W_2^{ff} \in \mathbb{R}^{d_{e} \times d_{ff}}$, $b_1^{ff}\in \mathbb{R}^{d_{ff}}$and $ b_2^{ff} \in \mathbb{R}^{d_{e}} $.

Both the attention and the feedforward components are followed by a skip connection and layer normalization (\cite{Vaswani2017}). The final output is then the output of the set encoding block, i.e., a representation vector $h_i^e \in \mathbb{R}^{d_{e}}$ for each node/point of interest.

\subsubsection{Transformer with Recursion}
\label{sec:transformer_recursion}
We changed the self-attention of the set encoder of previous work (\cite{Deudon2018, Kool2019a} by making it recursive. In order to make the set encoder block recursive, we changed the self-attention inputs by changing Equation \ref{eq:self_attention_key} to compute each key $k_i$ at iteration $l$ using the representation $h^{e}_i$  from the previous iteration: 
\begin{equation}
k_i = W_k h^{e, l-1}_i.
\end{equation}
We used $h^e_i$ for every self-attention layer of the set encoder block. This change forces the self-attention mechanism to be a function of the solution construction state at each iteration step. 
 
\subsubsection{Graph Masked Self-Attention}
\label{sec:graph}

In the OPTW, as in many other routing problems, at each iteration step $l$ of the iterative process of building a solution, the set of admissible locations, $\mathcal{A}^l( \mathrm{v}^l_{\ast}, t^l)$, may change significantly. At each iteration step, a different graph structure can be induced over the various elements of the admissible set of locations, $\mathcal{A}^l$, leveraging the computation of self-attention during the set encoding (\cite{Velickovic2018}). One straightforward approach to building such graph is to assume the set of admissible locations as a complete graph and use this graph to perform masked self-attention (see "complete graph" in Supplementary materials \ref{app:notweaks}). Previous studies (\cite{Kool2019a}) have referenced graph attention, however they typically use complete graph representations over the set of admissible nodes and do not take effective advantage of the dynamic graph structure that can be induced. In the next paragraph we explain how we refine that graph and clarify what we mean by masked self-attention.

We define a more refined graph structure by considering that the node or point of interest $\mathrm{v}_i$ is connected to node $\mathrm{v}_j$ if $[\mathrm{v}^l_{\ast},\mathrm{v}_i,\mathrm{v}_j,\mathrm{v}_n]$ is a feasible sequence, where $\mathrm{v}^l_{\ast}$ is the current node and $\mathrm{v}_n$ is the ending location. This is essentially a one-step lookahead search. At every iteration step $l$, we determine the adjacency matrix $Ad^l$ of the graph. When performing self-attention, the scores $u_{ij}$ are computed using Equation \ref{eq:simscore} only if $Ad_{ij}^l=1$, and are set to zero otherwise (i.e. are masked).  By masking not only the immediate non-admissible nodes, but also one-step lookahead paths, it is reasonable to expect better representation vectors for each node or point of interest.

Note that using graph masked self-attention and dynamic features requires updating the set representation at each iteration step. Performing set encoding only at the first step/iteration would make the model ignore these relevant pieces of information.
 \subsection{Sequence Encoding}

The main objective of this functional block is to compute a vector representation of the subsequence that is being built iteratively. We denote this vector by $ h^{d, l} $, with $ h^{d, l} \in \mathbb{R}^{d_{d}}$.

We call this block the "sequence encoding" block. The name might generate some confusion since, in NLP-translation/seq-2-seq applications where the Pointer Network architecture originated, this block is typically named the” decoding” block as it is translating/decoding into the target language. For the OPTW, and other combinatorial optimization problems, "sequence encoding" seems a more suitable name considering its role in the overall architecture. 

As in \cite{Vinyals2015} we chose a Long Short Term Memory (LSTM) (\cite{LSTM}) for sequence encoding. Since we compute a new set encoding at each iteration $l$, together with the hidden state vector of the previous step, $ h^{d, l-1}$ we can give as input to the LSTM the vector representation of the current point of interest computed by the set encoder in the current iteration step, $ h^{e, l}_{i} $. By relying on an up-to-date iterative encoding of each point of interest, we expect to obtain a better encoding of the sequence built up to step $l$.

\subsection{Pointing Mechanism}

On top of the set and sequence encoder blocks, lies the pointing mechanism (see Figure \ref{Pointnet_fig}). At each iteration step $l$, it takes as input both the set encoding of each node $\lbrace h^{e, l}_1,  h^{e, l}_2,\cdots,  h^{e, l}_n  \rbrace$ and the latest hidden state $ h^{d, l}$ of the sequence encoder's LSTM, and computes a probability distribution over the set of points of interest. This probability vector can then be use for selecting of the point of interest $\mathrm{v}^{l}$ of the tour.

To get to this probability vector we start by computing an importance score for each element of the input set using additive attention (\cite{Bahdanau2014})
 
\begin{equation}
u^l_{j} = \begin{cases}
    w^T \tanh (W_1 h^{e, l}_j + W_2 h^{d, l} ), \quad \forall j : \mathrm{v}_j \in \mathcal{A}^{l-1} \\
    -\infty , \quad \text{otherwise}
\end{cases}.
\end{equation}
where $w \in \mathbb{R}^{d_{h}}$ and the matrices $W_1  \in \mathbb{R}^{d_h \times d_e}$ and  $W_2  \in \mathbb{R}^{d_h \times d_d}$ are learnable parameters. Points of interest that cannot be visited, i.e. $\mathrm{v}_j \not\in \mathcal{A}^{l-1}$, are masked (\cite{Bello2016, Deudon2018, Kool2019a}). In order to control the range of the logits, $u_j^l$ is subsequently transformed by 
\begin{equation}
\tilde{u}^l_{j} = \begin{cases}
    C \tanh (u_{j}^l), \quad \forall j : \mathrm{v}_j \in \mathcal{A}^{l-1} \\
    -\infty , \quad \text{otherwise}
\end{cases}.
\label{eq:logits}
\end{equation}

where $C$  a hyper-parameter (\cite{Bello2016}). Finally we obtain the probability distribution over the points of interests after a softmax normalization of $\tilde{u}^l$

\begin{equation}
p_l(\mathrm{v}_j^{l}| A^{l-1}, S^{l-1}) =  \dfrac{\exp(\tilde{u}^l_{j})}{\sum_{j^{\prime}} \exp(\tilde{u}^l_{j{\prime}})}.
\end{equation}
Using this probability distribution, we can choose the next point of interest to be visited and after its selection, the iterative loop proceeds until a stopping condition (i.e. reaching $\mathrm{v}_n$) is met.

\section{Methods}
\label{sec:methods}

\subsection{Benchmark Instances}
\label{subsec:benchmark_instances}

We use three groups of benchmark instances: \textit{Solomon}, \textit{Cordeau} and \textit{Gavalas}. All these instances have best-known results documented in recent publications (\cite{KARABULUT2020106109, Gavalas2019}) and represent a diverse set, with different distributions of points of interest (Figs.  \ref{fig:stats_location_solomon}, \ref{fig:stats_location_cordeau} and \ref{fig:stats_location_gavalas}), scores and duration of visit (Figs. \ref{fig:stats_visit_dur_vs_score_solomon}, \ref{fig:stats_visit_dur_vs_score_cordeau} and \ref{fig:stats_visit_dur_vs_score_gavalas}) and schedules (Figs. \ref{fig:stats_schedules_solomon}, \ref{fig:stats_schedules_cordeau} and \ref{fig:stats_schedules_gavalas}). 

The Solomon and Cordeau instances are groups of established benchmarks in the OPTW literature (\cite{Vansteenwegen2011}). The Solomon group was originally adapted from a set of vehicle routing problems with time windows and is composed of $56$ instances, all with $100$ points of interest. The Cordeau group, includes $20$ instances, ranging from $48$ to $288$ points of interest.

The Gavalas instances (\cite{Gavalas2019}) were designed to be more representative of real TTDP problems. In particular, their times lay within a $24$-hour ($1440$ minutes) time range and they were generated with the assumption that a tourist will attach higher value, on average, to a point of interest that requires more time to visit, i.e., they all have a correlation between the scores and the duration of the visit of each point of interest (see Figure \ref{fig:stats_visit_dur_vs_score_gavalas}). The Gavalas group includes instances for the more general problem of Team OPTW, which can be framed as a tourist problem if we allow the tourist to use more than one day to visit the instance-region. Here, since we are only addressing the OPTW (and not the TOPTW), we consider only the $33$ Gavalas instances for which the tourist has a single day available. These instances range from from $100$ to $200$ points of interest (see Section \ref{app:instance_stats} in Supplementary materials and \cite{ OPbook} for further characterization of all benchmark instances).

The distances between locations are rounded to the first decimal place for the Solomon instances and to the second decimal place for the Cordeau and Gavalas instances. For most of our analysis, we focus on a subset of each of the groups, which we dubbed sub-Cordeau (8 instances), sub-Solomon (12 instances) and sub-Gavalas (8 instances), making it a total of 28 instances (see Section \ref{subsec:experiments}).

\subsection{Generated Instances}
\label{subsec:generated_instances}

Some of the parameters of an instance are tourist dependent and some others are invariant across tourists. We can look at each instance as the representation of a given tourist visiting a given region. Each tourist has a particular taste (scores), and preferences regarding starting locations and starting and ending times. Here we assume that the coordinates of the points of interest, the duration of the visit and the opening and closing times are instance-region parameters and remain unchanged regardless of the tourists. 

Having a model for a particular instance-region means having a model that works for all tourists that might visit that region. For that reason, for each benchmark instance we generate new tourists, i.e., new instances with the same instance-region parameters but different tourist parameters. We use these generated instances for training, monitoring and validation (see Sections \ref{sec:train} and \ref{sec:inf}). The choice of hyper-parameters for sampling was not done to specifically optimize generalization for the benchmark tourist-instance, but instead to generalize a reasonable range of tourist’s parameters while maintaining the same hyper-parameters across groups of instances.  

\subsubsection*{Generate Starting (and Ending) Location}
We sample uniformly a new starting location from the $[0,100]\times [0,100]$ square for Solomon and Gavalas instances and from the $[-100,100]\times [-100,100]$ square for Cordeau instances. In all the benchmark instances, the tour ends in the same position as it starts, and we assume the same for the generated instances.

\subsubsection*{Generate Starting and Ending Times}
Let us name $T_{start}^b$ and $T_{end}^b$ the starting time of the tour and the upper-bound on the tour's end time, respectively, for the tourist in the benchmark instance. We want to sample new times  $T_{start}^g$ and $T_{end}^g$ for a new generated tourist. We chose this sampling empirically because it seems realistic enough and works for the 3 groups of benchmark instances.  Most importantly we wanted it to span a range that is diverse enough and that would include the $T_{start}^b$ and $T_{end}^b$. 
    
\begin{figure}[H]
\includegraphics[width=0.95\textwidth]{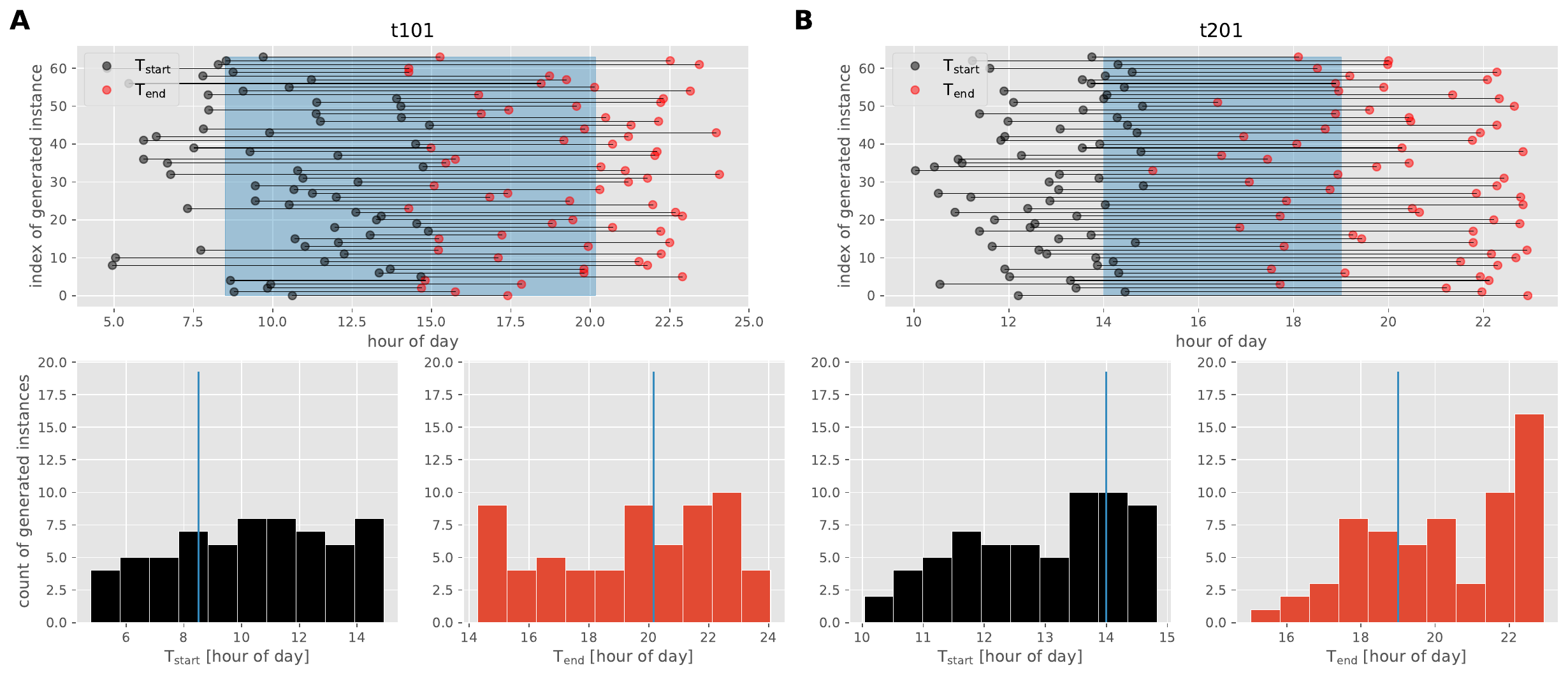}
\centering
\caption{Starting times ($T_{start}$, black) and upper-bounds on ending time ($T_{end}$, red) of 64 generated instances (tourists) for the instance-regions $t101$ (\textbf{A}) and $t201$ (\textbf{B}) of the Gavalas group. Blue shaded area (top panels) indicates the time period between $T_{start}$ and $T_{end}$ for the original benchmark instance. Vertical blue lines in the bottom panels indicate $T_{start}$ and $T_{end}$ of the original benchmark instance. Histograms in the bottom panels correspond to the data shown in the upper panels.}
\label{fig:sampling_time_gavalas}
\end{figure}

We start by defining the duration of a day for a benchmark-instance as the maximum time stamp of that instance: $T_{day}^b := \max\{T_{end}^b, \max_{i=1,..n}\{c_i\}\}$.
Then we normalize all times so that the duration of a day is 24 (as in 24 hours).
Finally we sample $T_{start}^g$ as \begin{equation*}
    T_{start}^g \sim \mathcal{U}(T_{start}^b-4, ub)
    \end{equation*}
    where
    \begin{equation*}
    ub = \min\{15, T_{end}^b+4\},
    \end{equation*}
i.e. the tour starts  4 "hours" at most before the benchmark-tourist, and before time 15 (3pm) unless the tourist in the benchmark-instance ends tour 4 "hours" before 3pm, in which case $T_{end}^b+4$ is the upper bound $ub$.
We sample $T_{end}^g$ uniformly:   
\begin{equation*}
T_{end}^g \sim \mathcal{U}(lb, T_{end}^b+4)
\end{equation*} where    \begin{equation*}
lb = \max\{12, T_{start}^g+4\},
\end{equation*}    i.e. $T_{end}^g$ is at most 4 "hours" after the benchmark-tourist, not before ”noon” and there also must be at least 4 hours left for the tour. After the sampling, we divide by 24 and multiply back by the duration of the day for the benchmark-instance and round it to the closest integer (see Figure \ref{fig:sampling_time_gavalas}).

Note that $T_{max} = \max\{ T_{day}^b,  T_{end}^b+4\}$ is an upper-bound on the max time stamp for all generated instances. We use this instance-region specific value to normalize the time input features. 
 
\subsubsection*{Generate Scores}
The Gavalas instances were created to reflect essential characteristics of a realistic TTDP. In particular, the authors assumed that the scores that a tourist gives to each point of interest is correlated with the time it takes to visit that point of interest. The authors do not specify how this correlation was generated (\cite{Gavalas2019}) but, in an attempt to improve learning, our aim was to generate instances with the same property. For that reason, when sampling scores we tried two different sampling schemes for the Gavalas instances: a uniform sampling scheme and a non-uniform or correlated one.

 In the uniform scheme the scores are uniformly sampled from $[1, 1.1\times S^b_{\max}]$ where $S^b_{\max}$ is the maximum score for the benchmark instance. This is the scheme used for the Solomon and Cordeau generated instances. In the correlated sampling scheme, the scores are generated according to:
\begin{equation}
\label{eq:corr_score_sample}
s_i \sim \mathcal{N}(S^b_{\max} \frac{d_i}{d_{max}} ,\,100)\,
\end{equation}and clipped to be in the $[1, 1.1\times S^b_{\max}]$ interval, where $d_{max}=\max_{j=1,...,n}\{d_j\}$ is the maximum duration of visit across all points of interest of that instance. Note that $1.1\times S^b_{\max}$ is an upper-bound on the scores of the generated instances, which we will use as a normalization factor for score-related features. For visualizations of the correlation between score and duration of visit for every benchmark instance see Figures  \ref{fig:stats_visit_dur_vs_score_solomon}, \ref{fig:stats_visit_dur_vs_score_cordeau} and \ref{fig:stats_visit_dur_vs_score_gavalas}.

These generated instance-tourists are used to train the model: during training they are generated on the fly for the given instance-region the model is being trained for (see Section \ref{sec:train}). For each instance of the three benchmark groups, we have also created a fixed validation set consisting of 64 generated instances. These fixed validation sets are used to monitor each model’s training and evaluate its performance.

\subsection{Input Features}

The first encoding block of the neural network has a set of feature vectors as input: one feature vector for each point of interest (see Section \ref{sec:set_encoding_block}). These features can be divided into 2 groups: the static features and the dynamic features.

The static features remain constant during the iterative process of building a solution. Here, we use 7 straightforward features retrieved directly from the instance data, namely the point of interest's Euclidean coordinates, visiting duration, opening and closing time and score, and also the upper bound on the finishing time of the trip, $T_{end}$. The Euclidean coordinates are mapped using min-max scaling into unit square $[-1,1]\times [-1,1]$. 
The scores are normalized by $1.1\times S^b_{\max}$ and all the time features are normalized by $T_{\max}$. Note that these normalization constants are the same across tourists for a same instance-region.

The dynamic features can change at every iteration step. We designed 8 features that are functions of the current time, $t^l$, and current point of interest, $v^l_{\ast}$: for each point of interest, the time left until the opening time and time left until closing time; the fraction of time elapsed since the $T_{start}$; the fraction of the time left to $T_{end}$. To these features we add the same set of 4 features but assuming we have traveled from the current position to the point of interest that the feature corresponds to (remember that there is one feature vector for each point of interest), i.e., adding to $t^l$, the travel time from $v^l_{\ast}$ to $v_{i}$. All the dynamic time features are normalized by the maximum time available for the tourist to perform the tour, i.e.  $T_{end} - T_{start}$. Thus, contrary to the static time features, this normalization constant is tourist dependent.

\subsection{Training}
\label{sec:train}

We train the neural network model using reinforcement learning. 

\subsubsection*{Objective Function}
For a given instance $\phi$, the Pointer Network model, with parameterization $\theta$ defines, at each iteration step, a stochastic policy from which the next visiting location can be sampled. We can use this policy iteratively until the ending location is reached and obtain a solution/route $ S = (\pi_1, \ldots, \pi_m)$. For simplicity we can represent this by $S \sim p_{\theta}( \cdot | \phi) $. 

The total route probability can be determined using the chain rule of probability: 
$$
p_{\theta}(S | \phi) = \prod_{l=1}^m p_{\theta}(\pi_l| \phi, \mathcal{A}^{l-1}, S^{l-1}).
$$
We want to determine $\theta$ such that we can sample high total score solutions from its policy with high probability. 
 
For a sample solution $S$, the total score is given by $R(S) = \sum_{l=1}^m r(\pi_l)$ and our objective function is defined as the expected total score, that for instance $\phi$ is given by:
$$J\left(  \mathbf{\theta}| \phi \right) = \mathbb{E}_{S \sim p_{\theta}( \cdot | \phi)}[ R(S)]$$

\subsubsection*{REINFORCE Algorithm}
We maximize $J(\theta)$ using gradient ascent, and resort to REINFORCE algorithm (\cite{Williams1992}) to estimate the gradients (see Algorithm \ref{alg:rein}).

\begin{algorithm}[ht]
\KwIn{training set $\Phi$, batch size $B$}
  
Initialize network parameters $\theta$ \;

\While{ training not finished}{
sample instance $\phi$ from $\Phi$ \;

\For{every $b \in \{1,\ldots ,B\}$}{
$l  \leftarrow 1$

\While{terminal node not reached}{

sample $\pi_l$ from $ p_{\theta}(\cdot | \phi, \mathcal{A}_b^{l-1}, S_b^{l-1})$ \;

$ l \leftarrow l+1 $ \;

}
}

$\overline{R}  \leftarrow \frac{1}{B}\sum_{b=1}^B R(S_b)$ \;

$g_{\theta} \leftarrow  -\frac{1}{B} \sum_{b=1}^{B} \left(  R(S_b) - \overline{R} \right)\mathbf{\nabla}_{\mathbf{\theta}}\log
 p_{\mathbf{\theta}}\left(  S_b \left\vert \phi \right.
\right)$\;

update $\theta$ using $g_{\theta}$\;

}
\caption{REINFORCE algorithm}
\label{alg:rein}
\end{algorithm}
 
Typically, when using the REINFORCE algorithm, each batch is composed by $B$ different (generated) instances. In our approach, however, each batch comprises a single generated instance and thus $B$ independent solutions are sampled for that same single tourist-instance. This allows a straightforward baseline estimation, thus avoiding the need to maintain a moving average estimate of the baseline, obtain an estimate of it through a greedy roll-out computation,(\cite{Kool2019a}), or having to rely on a critic neural network, (\cite{Bello2016, Vaswani2017}). In concrete terms, we determine the batch average total score, $\overline{R} = \frac{1}{B}\sum_{b=1}^B R(S_b) $, and use it as our baseline to reduce gradient estimation variance. This approach is a particular case of the multiple sampling with replacement method (\cite{Kool2019b} with $k=1$) where the number of solution samples equals to the batch size. 

\subsubsection*{Other Training Details}
We train all models with the Adam optimizer (\cite{Kingma2015AdamAM}) and use a batch size of $B=32$. We use 500,000 epochs for models trained from scratch and 50,000 epochs for the models that are fine-tuned (see \ref{subsec:experiments}). For models trained from scratch all parameters are initialized with the Xavier uniform initialization (\cite{Glorot2010}), and an initial learning rate of $10^{-4}$ decaying every 5000 steps by a factor of $0.96$, until a minimum of $10^{-5}$. For fine-tuning we use a fixed learning rate of $10^{-5}$. Both the first hidden and the first cell state vectors of the sequence encoder's LSTM are treated as learnable parameters and initialized from $\mathcal{U}[-\frac{1}{\sqrt{d_h}}; \frac{1}{\sqrt{d_h}}]$.

We set the hyper-parameter $C=10$, in Equation \ref{eq:logits}, as suggested in previous literature (\cite{Bello2016}). We tried to set it to be a learnable parameter without much success in the few attempts and confirmed empirically the quality of the suggested value.

We use 2 transformer blocks with 8 heads in the multi-head self-attention. We use $d_h =d_e = d_d = 128$ and $d_{ff} = 256$.

\subsection{Inference}
\label{sec:inf}

Inference is the process of building a solution for a particular instance-tourist from the model (or network policy) trained in that instance-region.
We use four different solution construction strategies, namely: sampling, greedy, beam search, and active search. We use sampling during training in the REINFORCE algorithm and greedy search to monitor the evolution of learning. In a sampling strategy, for each step, we sample from the probability vector that the model outputs. In a greedy strategy, for each time step, we select the node/point of interest for which the model gives the highest probability, that is $\pi_l = \argmax_{\mathrm{v}_i\in \mathcal{A}^{l-1}}p_{\theta}(\cdot |\phi,\mathcal{A}^{l-1}, S^{l-1})$.

\subsubsection*{Beam Search}
Beam search is a search heuristic that approximately maximizes the total route probability under the network policy $p_{\theta}(\cdot|\phi)$. It keeps a record, at each time step, a list of the $n_b$ most promising partial solutions sequences. From these candidates, beam search considers all admissible next visiting locations, and selects the best top-$n_b$ of them. When every beam reaches the final location, the solution with higher score is selected. When $n_b=1$, beam-search heuristic reduces to a simple greedy search. For beam search inference we show performance for $n_b$ ranging from 1 up to a maximum of 128 and finally consider performance and processing times for a maximum of 128 beams. Note that the number of beams cannot be higher than the number of nodes in the instance.

\subsubsection*{Active Search (with Beam Search)}
 Active Search (\cite{Bello2016}) optimizes the network policy for the given tourist-region instance, i.e., given an instance $\phi$ we can retrain the model's weights directly on that instance. The vanilla version uses greedy inference, here we use beam search (see Algorithm \ref{alg:Actvearch}).

\begin{algorithm}[ht]
\caption{Active Search algorithm} \label{alg:Actvearch}
\KwIn{instance $\phi$, network parameters $\theta$, batch size $B$}

\While{$epoch \leq epoch_{max}$}{

\For{all $b \in \{1,\ldots ,B\}$}{
$l  \leftarrow 1$

\While{terminal node not reached}{

sample $\pi_l$ from $ p_{\theta}(\cdot | \phi, \mathcal{A}_b^{l-1}, S_b^{l-1})$ \;

$l \leftarrow l+1 $\;

}
}

$\overline{R}  \leftarrow \frac{1}{B}\sum_{b=1}^B R(S_b)$ \;

$g_{\theta} \leftarrow  -\frac{1}{B} \sum_{b=1}^{B} \left(  R(S_b) - \overline{R} \right)\mathbf{\nabla}_{\mathbf{\theta}}\log
 p_{\mathbf{\theta}}\left(  S_b \left\vert \phi \right.
\right)$\;

update $\theta$ using $g_{\theta}$\;

}

$S  \leftarrow \text{Beam Search}( \phi, \theta)$\; 
\end{algorithm}

Even though it is considered an inference method, active search can be seen as a way to achieve training/fine-tuning of the region-model for a specific tourist. For active search, we use 128 epochs and a maximum of 128 beams for the beam search that follows active search.

\subsection{Experiments}
\label{subsec:experiments}

In addition to exploring different kinds of inference, we also explore different training schemes. All the experiments conducted allowed us to obtain results for at least a subset of 28 benchmark instances (or 3 subsets, 8 of Cordeau, 8 of Gavalas and 12 of Solomon). The instances of these subsets are: Solomon: c101, c102, r101, r102, rc101, rc102, c201, c202, r201, r202, rc201, rc202; Cordeau: pr01, pr02, pr03, pr04, pr11, pr12, pr13, pr14; Gavalas: t101, t105, t114, t117, t201, t202, t203, t204.  We chose these subsets because we wanted to have results for at least 8 instances of each group and at the same time have a balanced representation of the diversity within each group. In all models/experiments we use beam search for inference with up to a maximum of 128 beams.

\subsubsection*{Models Trained from Scratch}
For the main result, we train models from scratch for 500,000 epochs for each of the subset of 28 instances. Then, we look at performance with ("\textit{model+as}") and without ("\textit{model}") 128 epochs of active search. We also look at the performance of "model" at 50,000 epochs of training: we denote this model by "\textit{st}".

\subsubsection*{Transfer Learning}
For each instance group, we also use the instances in the subset of 28 instances as leave-out sets for evaluating the feasibility of using transfer learning: we train a global model on all instances except for the instances in the leave-out set and then evaluate its performance regarding those left-out instances. We called this model "\textit{tl}".  Furthermore, we evaluate the performance of fine-tuning that global model on each of those left-out instances of the subset. Fine-tuning happens for $10\%$ of regular training, i.e. 50,000 epochs. We called these fine-tuned models "\textit{ft+tl}". While training a global model, i.e., a model in several instances at the same time, the instance chosen for each batch in the REINFORCE algorithm is generated from one randomly chosen benchmark instance of the training set of instances.

\subsubsection*{General global Model}
Finally, we also train a global model on all instances simultaneously and then fine-tune it for each of the instances. This was done to investigate the feasibility of warming up a general model when we have a lot of different instance-regions to train.

All experiments were performed on a 12 core CPU at 3.5GHz (AMD Threadripper 1920X) with 64 GB of RAM and an Nvidia GeForce GTX 1080 Ti GPU.

\subsection{Metrics and Statistical analysis}
To quantify performance, we use total score of the solution, i.e., the sum of the scores/rewards. For each generated instance group, the reported score is the mean value of the average score across the 64 instances of the validation sets within that group.

We compare our model's performance to the Iterate Local Search (ILS) heuristic (\cite{Vansteenwegen2009}), as it presents one of the best trade-offs between solution quality and execution time. We also report the score gap to the best-known solution in the benchmark instances. We define score gap to a baseline model as $$ \frac{score_{baseline}-score_{model}}{score_{baseline}}\times 100$$
We use bootstrapping to obtain a 95\% confidence intervals for gap to ILS and gap to best known. We sample with replacement the benchmark instances or instant-regions 10000 times. For determining the p-values in pairwise comparisons of scores we use the non-parametric one-sided Wilcoxon signed-rank test.

\section{Results}

We investigate the performance of an attentive deep reinforcement learning approach to solving the Orienteering Problem with Time Windows (OPTW). We use a Pointer Network architecture and the REINFORCE algorithm to estimate gradients during training. We evaluate the performance of greedy inference as well as that of beam search and active search. We address the problem with the Tourist Trip Design Problem (TTDP) application in mind, but our model’s performance and inference times do not limit it to that application. We consider 3 groups of benchmark tourist-instance-regions (the “benchmark” instances) and generate new tourists for each of the instance-regions (the “generated” instances) in a way that mimics variations expected in the TTDP applications (i.e., the variability across tourists). In concrete terms, from each benchmark instance we sample new instances (new tourists for that instance’s region) with different route starting positions, different route starting and ending times, and different tourist-specific scores/preferences for each point of interest. Then, we compare performance with a well-established heuristic, the Iterated Local Search (ILS) algorithm, on both the benchmark and generated instances, and quantify inference speeds. 
Finally, we explore the practicality and performance of transfer learning and fine-tuning training schemes. 

 \subsection*{The model is able to learn and achieve production-level performance}
 \label{results:full}
 
\begin{figure}[!htb]
\includegraphics[width=0.95\textwidth]{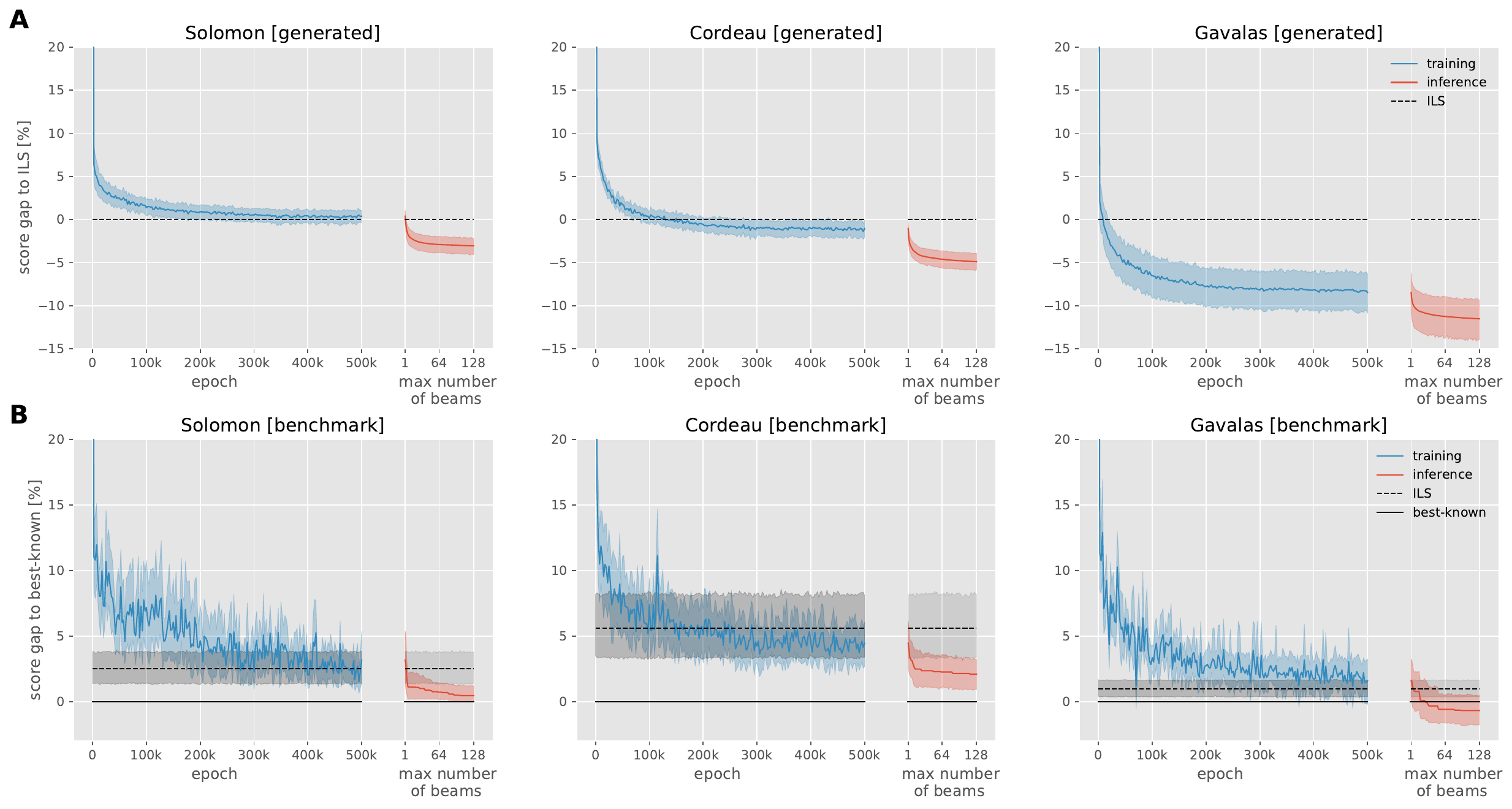}
\centering
\caption{Evolution of model performance during training using greedy inference (blue) and post-training model performance using beam search for inference (red) with up to a maximum number of 128 beams, in both benchmark and generated instances for the three sub-groups of instance-regions (n=12 for Solomon, n=8 Cordeau and n=8 for Gavalas). \textbf{A.} Average score gap to ILS on the generated instances (n=64 generated instances per instance-region). \textbf{B.} Average score gap to best-known and ILS-score gap to best-known on the benchmark instances. Inference during training (blue) uses greedy inference. Shaded area represents 90\% confidence intervals (bootstrap instance-regions).}
\label{fig:all_inst_val}
\end{figure}

First we wanted to know if the model can learn a particular instance-region when trained from scratch. We evaluate training on a subset 28 instances (n=12 for Solomon, n=8 for Cordeau and n=8 for Gavalas, see Section \ref{sec:methods}.  Methods for details). We find that the model is able to learn for each set of instances on both the generated instances (Fig. \ref{fig:all_inst_val}A blue) and also that it generalizes to the benchmark instances (Fig. \ref{fig:all_inst_val}B blue) during training. At the end of training the performance (using greedy inference) is already in line with ILS. We find that beam search significantly improves performance (Fig. \ref{fig:all_inst_val}A and B red) and that it is generally able to significantly outperform ILS (Table \ref{table:all_inst_real}) with inference times below half a second.  In fact, we find that already with only 10\% of training (50,000 epochs) the model already has higher scores than ILS in the generated instances ($p=2.6E-06$, one-sided Wilcoxon signed-rank  test, see "ILS$\to$st" in Figure \ref{fig:transfer_learning_fig_pvalues}). Our model is able to learn and achieve competitive production level performance in both scores and execution times. 

\begin{table}[!htb]
\renewcommand{\arraystretch}{1.5} 
\centering
\resizebox{.50\width}{!}{\rowcolors{2}{black!5}{white}
\begin{tabular}{cccccccccc}
\toprule 
\textbf{instance group} & \textbf{best-known (bk)} & \textbf{ILS} & \textbf{model score} & \textbf{gap to ILS} & \textbf{gap to ILS [95\%CI]} & \textbf{gap to bk} & \textbf{gap to bk [95\%CI]} & \textbf{time [s]} & \textbf{time [95\%CI]} \\
\midrule 
Solomon [generated] & - & 429.99 & 438.72 & \textcolor{dgreen}{-3.06\%} & [-4.27\% , -1.98\%] & - & - & 0.155 & [0.099 , 0.212] \\
Cordeau [generated] & - & 253.69 & 266.61 & \textcolor{dgreen}{-4.90\%} & [-6.12\% , -3.77\%] & - & - & 0.245 & [0.148 , 0.354] \\
Gavalas [generated] & - & 299.99 & 333.80 & \textcolor{dgreen}{-11.52\%} & [-14.44\% , -8.85\%] & - & - & 0.349 & [0.258 , 0.464] \\
Solomon [benchmark] & 575.58 & 558.83 & 571.33 & \textcolor{dgreen}{-2.19\%} & [-3.81\% , -0.85\%] & \textcolor{dred}{0.46\%} & [0.02\% , 1.27\%] & 0.301 & [0.201, 0.406] \\
Cordeau [benchmark] & 428.00 & 401.62 & 418.25 & \textcolor{dgreen}{-3.90\%} & [-7.34\% , -0.7\%] & \textcolor{dred}{2.10\%} & [0.75\% , 3.58\%] & 0.412 & [0.265, 0.573] \\
Gavalas [benchmark] & 310.75 & 307.50 & 313.88 & \textcolor{dgreen}{-1.70\%} & [-3.4\% , 0.15\%] & \textcolor{dgreen}{-0.67\%} & [-1.95\% , 0.71\%] & 0.344 & [0.214, 0.509] \\
\bottomrule 
\end{tabular}}
\caption{Average performance (average score, gap to ILS and gap to best-known) and inference times of beam search inference with a maximum number of 128 beams for models trained from scratch for 500,000 epochs on each of the instance-regions in the subset of instances (12 Solomon, 8 Cordeau and 8 Gavalas). The generated instances include n=64 generated instances for each of template benchmark instance-regions. The 95\% confidence intervals are computed using bootstrap (sampling instance-regions with replacement). For performance on each of the 28 individual benchmark instances and average performance for each individual instant-region see "model" in Tables \ref{table:benchmark_results} and  \ref{table:generated_results} in Supplementary materials.}
\label{table:all_inst_real}
\end{table}

 \subsection*{Active search inference improves performance at the cost of speed}

\begin{figure}[!htb]
 \includegraphics[width=0.95\textwidth]{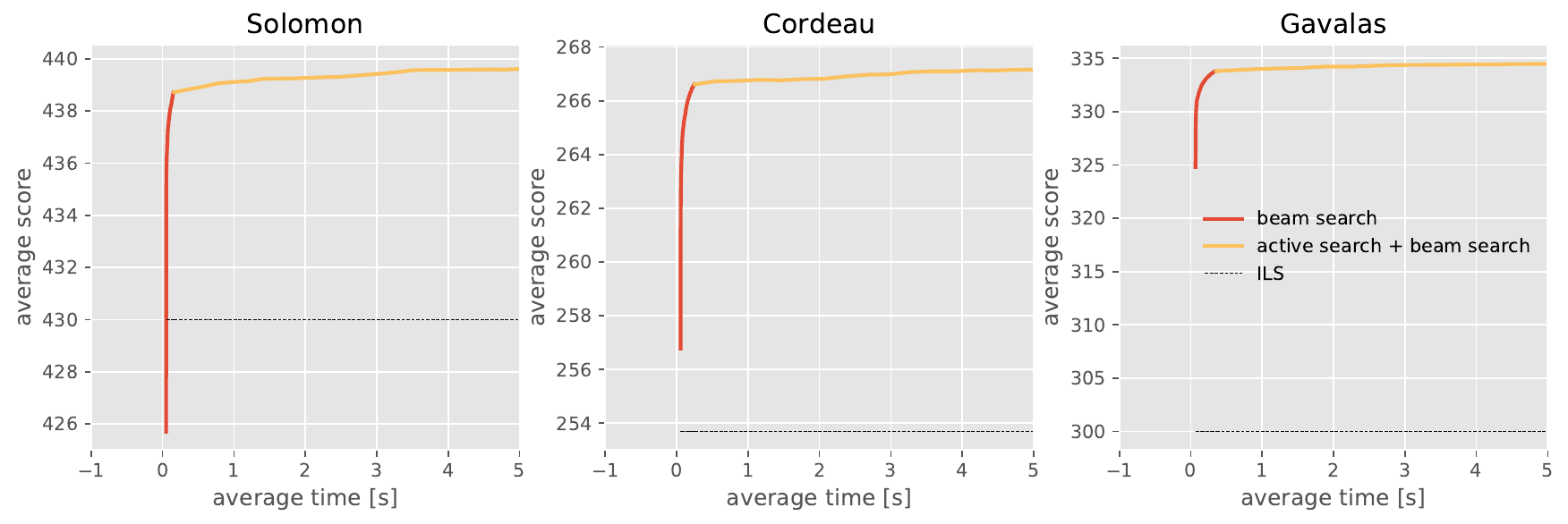} 
 \centering
 \caption{Average score in the generated instances as a function of inference time. Results are for 64 generated instances/tourists for each of the 28 benchmark template instance-regions in the subgroup of instances trained individually (12 Solomon, 8 Cordeau and 8 Gavalas). Scores are for inference using beam search changing the maximum number of beams from 1 to 128 (red) and for active search (yellow) with 1 and up to 128 epochs followed by beam search with a maximum of 128 beams. The x-axis is capped at 5 seconds for visualization purposes. See Table \ref{table:active_search} for inference times with 128 epochs of active search.}
 \label{fig:times}
 \end{figure}

Beam search inference produces good results in less than half a second. However, if the user is willing to wait a couple of minutes for a better solution, active search (\cite{Bello2016}, see Algorithm \ref{alg:Actvearch}) can be an appropriate inference solution. In active search the model is fine-tuned on for a particular tourist-region instance, improving its policy for solving the problem for that tourist. We apply active search to the models trained from scratch on the same subset of 28 instances, before applying beam search.
Aiming for realistic off-line computation times (under 2 minutes), we applied 128 epochs of active search for each benchmark and generated tourist-region instance. We found that indeed while it significantly slower (it is considerably slower than beam search (see Figure \ref{fig:times} and Table \ref{table:active_search}) it improves the scores ($p=1.6E-05$, one-sided Wilcoxon signed-rank  test; see "model$\to$model+as" in Figure \ref{fig:transfer_learning_fig_pvalues}, and Table \ref{table:active_search}). Active search is slower than beam search but can give a justified performance boost for situations where an immediate response is not necessary.

 \begin{table}[!htb] \renewcommand{\arraystretch}{1.5} 
 \centering 
 \resizebox{.50\width}{!}{\rowcolors{2}{black!5}{white}
\begin{tabular}{cccccccccc}
\toprule 
\textbf{instance group} & \textbf{best-known (bk)} & \textbf{ILS} & \textbf{model score} & \textbf{gap to ILS} & \textbf{gap to ILS [95\%CI]} & \textbf{gap to bk} & \textbf{gap to bk [95\%CI]} & \textbf{time [s]} & \textbf{time [95\%CI]} \\
\midrule 
Solomon [generated] & - & 429.99 & 439.76 & \textcolor{dgreen}{-3.25\%} & [-4.35\% , -2.26\%] & - & - & 27.069 & [19.3 , 35.2] \\
Cordeau [generated] & - & 253.69 & 267.32 & \textcolor{dgreen}{-5.14\%} & [-6.33\% , -4.06\%] & - & - & 39.252 & [25.1 , 54.8] \\
Gavalas [generated] & - & 299.99 & 334.50 & \textcolor{dgreen}{-11.73\%} & [-14.62\% , -9.16\%] & - & - & 49.553 & [37.5 , 65.1] \\
Solomon [benchmark] & 575.58 & 558.83 & 571.00 & \textcolor{dgreen}{-2.16\%} & [-3.77\% , -0.82\%] & \textcolor{dred}{0.50\%} & [0.03\% , 1.26\%] & 47.682 & [33.4, 62.5] \\
Cordeau [benchmark] & 428.00 & 401.62 & 420.25 & \textcolor{dgreen}{-4.38\%} & [-7.74\% , -1.5\%] & \textcolor{dred}{1.64\%} & [0.65\% , 2.76\%] & 61.512 & [41.9, 84.1] \\
Gavalas [benchmark] & 310.75 & 307.50 & 316.12 & \textcolor{dgreen}{-2.71\%} & [-4.04\% , -1.21\%] & \textcolor{dgreen}{-1.67\%} & [-2.61\% , -0.64\%] & 48.184 & [31.6, 68.5] \\
\bottomrule 
\end{tabular}} 
 \caption{Average performance of inference using active search with 128 epochs followed by beam search with a maximum number of 128 beams. Results are for the subset of instance-regions trained from scratch (n=12 for Solomon, n=8 Cordeau and n=8 for Gavalas). The generated instances include n=64 generated instances for each of template benchmark instance-regions. The 95\% confidence intervals are computed using bootstrap (sampling instance-regions with replacement). For performance on each of the 28 individual benchmark instances and average performance for each individual instant-region see "model+as" in Tables \ref{table:benchmark_results} and  \ref{table:generated_results} in Supplementary materials.} 
 \label{table:active_search} 
 \end{table}

 \subsection*{Transfer learning and fine-tuning}

Next, we investigate the possibility of speeding up training by fine-tuning a model trained on another set of relatively similar instance-regions. This could be relevant in practice, if there is some change to the current instance-region, e.g., opening times changes, new points of interest opened and others closed, etc., and we do not want to train the model from scratch. It could also happen in practice, if one wants to train for a new instance-region while potentially taking advantage of an already warmed-up model. We train on all instances of each of the 3 sets of instances while leaving out the same subset of instances we have considered so far (i.e., we leave out that same subset of 12 Solomon, 8 Cordeau, 8 Gavalas instance-regions and train on the remaining 44 Solomon, 12 Cordeau and 25 Gavalas instance-regions).

\begin{figure}[!htb]
\includegraphics[width=0.95\textwidth]{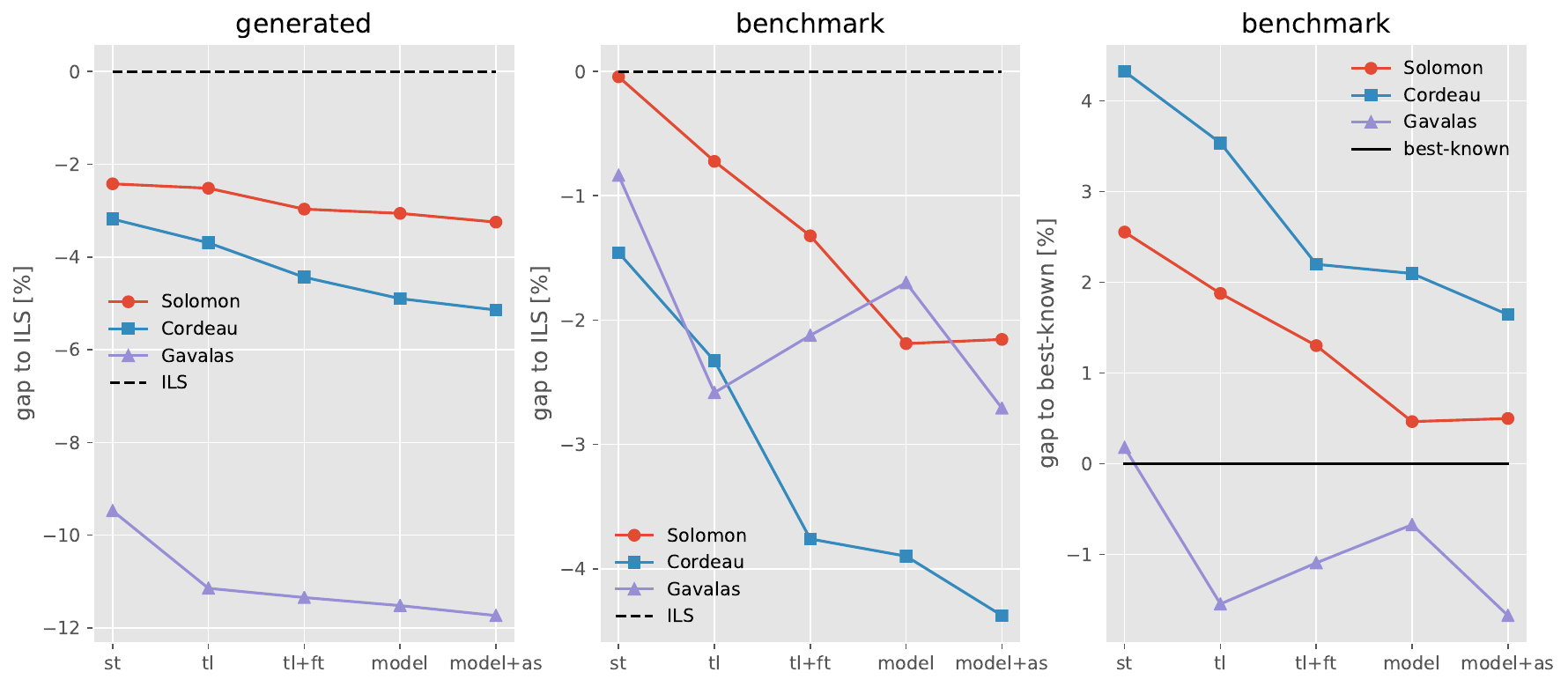}
\centering
\caption{Performance of the different training/inference strategies. Gap to ILS in generated (left panel) and benchmark (middle panel) instances, and gap to best-known in benchmark instances (right panel), within the subset of 28 instances (12 Solomon, 8 Cordeau and 9 Gavalas), for the different training or inference strategies: training for 50,000 epochs from scratch (\textit{st}); transfer learning of a model trained simultaneously during 500,000 epochs in all instances within each group except for those in the subset of 28 instances (\textit{tl}), fine-tuning the transfer learning model for each instance during 50,000 epochs (\textit{tl+ft}), training from scratch on each individual instance-region for 500,000 epochs (\textit{model}), "model" with added 128 epochs of active search inference (\textit{model+as}) before beam search. All models use beam search with a maximum of 128 beams. In the "generated" panel (left) we consider 64 fixed generated tourists for each instance-region, while each for benchmark instances there is a single tourist per instance. See Figure \ref{fig:transfer_learning_fig_pvalues} for more granularity and further analysis on the left panel data.}
\label{fig:transfer_learning_fig}
\end{figure}
\begin{figure}[!htb]
\includegraphics[width=0.65\textwidth]{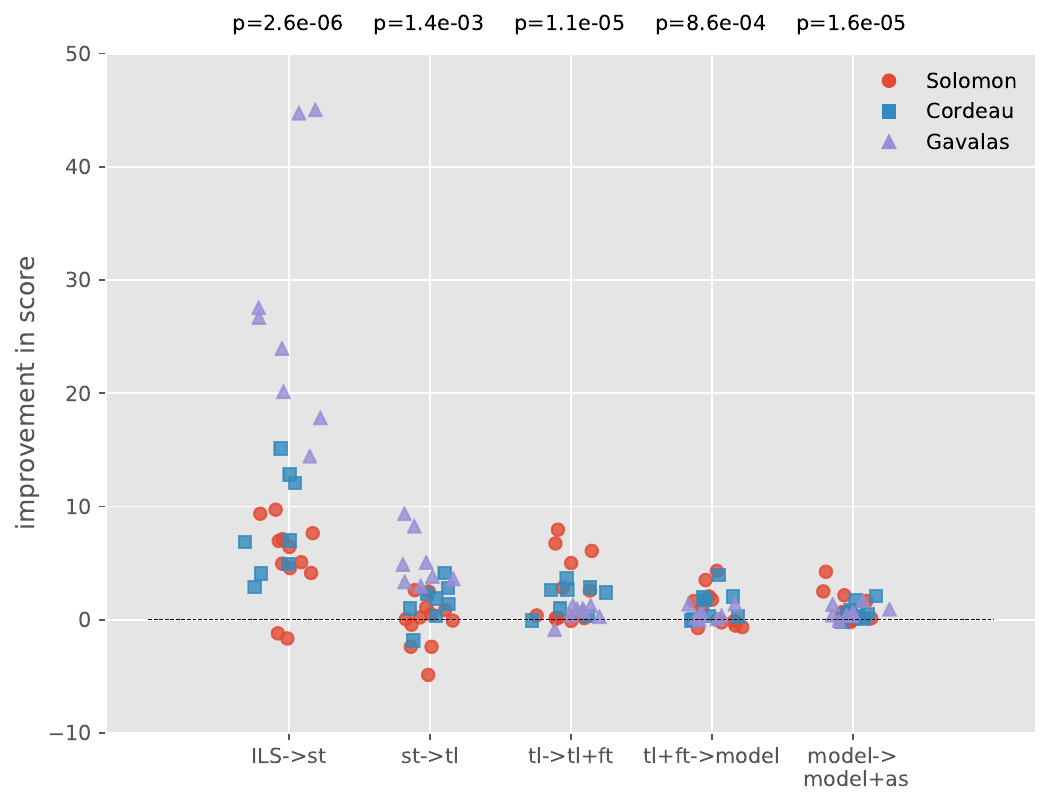}
\centering
\caption{Improvement in score in generated instances for different training/inference strategies. Each data point is the average of 64 generated tourists for each instance-region (see Figure \ref{fig:transfer_learning_fig} left panel).  Inference in all models uses beam search with up to a maximum of 128 beams. Training for 50,000 epochs from scratch (\textit{st}); transfer learning of a model trained simultaneously during 500,000 epochs in all instances within each group except for those in the subset of 28 instances (\textit{tl}), fine-tuning the tl model for each instance during 50,000 epochs (\textit{tl+ft}), training from scratch on each individual instance-region for 500,000 epochs (\textit{model}), "model" with added 128 epochs of active search inference (\textit{model+as}) before beam search. For example, "ILS$\to$st" represents the average score improvement from the ILS algorithm to the model trained for just 50,000 epochs across the 64 fixed generated instances for each instance-region in the subset of the 28 (12 Solomon, 8 Cordeau, 8 Gavalas). The p-values are from one-sided Wilcoxon signed-rank test.}
\label{fig:transfer_learning_fig_pvalues}
\end{figure}

We find out that with transfer learning we are able to reach good levels of performance in all subgroups (see Figure \ref{fig:transfer_learning_fig}). Even without any fine-tuning, it achieves higher scores than ILS ($p=2.9E-06$, one-sided Wilcoxon signed-rank test) and it is better than training on each specific instance from scratch for just 50,000 epochs ($p=1.4E-03$, one-sided Wilcoxon signed-rank test, see "st$\to$tl" in Figure \ref{fig:transfer_learning_fig_pvalues}). We find also that fine-tuning for 50,000 epochs further improves the transfer learned model ($p=1.1E-05$, one-sided Wilcoxon signed-rank test, see "tl$\to$tl+ft" in Figure \ref{fig:transfer_learning_fig_pvalues}) to performance levels closer to training from scratch even if still lower ($p=8.6E-04$, one-sided Wilcoxon signed-rank test,  see "tl+ft$\to$model" in Figure \ref{fig:transfer_learning_fig_pvalues}).

 \subsection*{Importance of representative generated instances}

\begin{figure}[!htbp]
\includegraphics[width=0.35\textwidth]{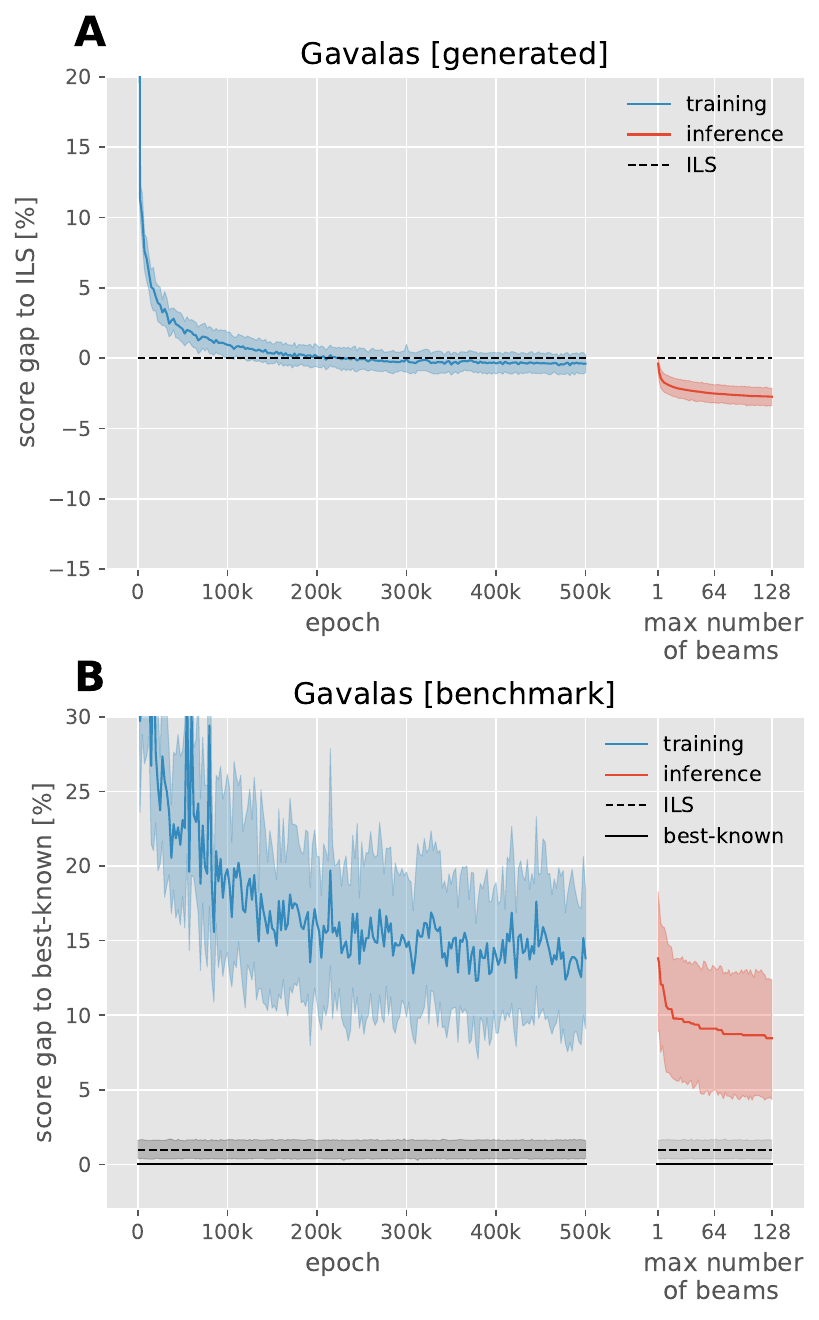}
\centering
\caption{Evolution of model performance during training using greedy inference (blue) and post-training model performance  using beam search for inference (red) with up to a maximum number of 128 beams, in both benchmark and generated instances for the three sub-set of n=8 Gavalas instance-regions using uniform sampling of rewards/scores (for Gavalas with correlated sampling see Figure \ref{fig:all_inst_val} right panels). Evolution of the mean validation score during training and model performance using Beam Search for inference. {A.} Average score gap to ILS on the generated instances (n=64 generated instances per instance-region). {B.} Average score gap to best-known and ILS-score gap to best-known on the benchmark instances. Inference during training (blue) uses greedy inference. Shaded area represents 90\% confidence intervals (bootstrap instance-regions).}
\label{fig:gavalas_uni}
\end{figure}

In generating new instances, we aimed at mimicking the kind of variability expected across tourists for a particular instance-region. We did this while trying to use the same hyperparameters across groups of instances (see Section \ref{subsec:generated_instances} in Methods). However, in the Gavalas set, we are given some information about the generative process (\cite{Gavalas2019}). In concrete terms, we know that the score of each point of interest is correlated with how long it takes to visit that point of interest. We used that information when generating new tourists for each instance-region of the Gavalas set. We found that ignoring that information leads to weak performances on the real instances (see Figure \ref{fig:gavalas_uni} B vs Figure \ref{fig:all_inst_val} B 3rd column). It also leads to weaker performances on the generated instances (see Figure \ref{fig:gavalas_uni} A vs Figure \ref{fig:all_inst_val} A 3rd column) even though it still manages to outperform the ILS algorithm for these. This probably happens because forcing that correlation reduces the dimensionality of the space of generated instances and therefore the model trains more efficiently. On the other hand, since the distribution of instances used during training is closer to the benchmark instances, the model generalizes better for that instance. Having a better generative process or sampling method improves the performance of the model.

 \subsection*{Fine-tune a global model for every instance}
 \label{subseb:global}

Finally, we explore another possible practical scenario in which the OPTW has to be solved for a good number of different instance-regions. If any cost, energy, or time restriction presents itself as an impediment, it may be wiser not to train a model from scratch for each instance-regions. Instead, we can take advantage of the fact that transfer-learning and fine-tuning produce good results, and in a similar way obtain a model for each instance-region in a reasonable amount of time. With this aim, we start by training one global model for each instance group (without leaving any instances out) and then fine-tune it for $10\%$ (50,000) of the epochs on each individual instance-region. By global we mean that it is simultaneously trained on all the instance-regions of that group. We thus obtain one fine-tuned model for each instance-region with considerable savings in resources. We find that both on the subset of 28 instances (Table \ref{table:sub_global}) as well as when looking at all instances (Table \ref{table:all_global}) this model and training scheme is able to outperform ILS.

\begin{table}[htbp]
\renewcommand{\arraystretch}{1.5} 
\centering
\resizebox{.55\width}{!}{\rowcolors{2}{black!5}{white}
\begin{tabular}{cccccccc}
\toprule 
\textbf{instance group} & \textbf{best-known (bk)} & \textbf{ILS} & \textbf{model score} & \textbf{gap to ILS} & \textbf{gap to ILS [95\%CI]} & \textbf{gap to bk} & \textbf{gap to bk [95\%CI]} \\
\midrule 
Solomon [generated] & - & 429.99 & 437.59 & \textcolor{dgreen}{-2.93\%} & [-4.19\% , -1.75\%] & - & - \\
Cordeau [generated] & - & 253.69 & 265.60 & \textcolor{dgreen}{-4.55\%} & [-5.56\% , -3.55\%] & - & - \\
Gavalas [generated] & - & 299.99 & 333.26 & \textcolor{dgreen}{-11.35\%} & [-14.38\% , -8.74\%] & - & - \\
Solomon [benchmark] & 575.58 & 558.83 & 566.08 & \textcolor{dgreen}{-1.41\%} & [-3.05\% , -0.17\%] & \textcolor{dred}{1.22\%} & [0.34\% , 2.29\%] \\
Cordeau [benchmark] & 428.00 & 401.62 & 417.62 & \textcolor{dgreen}{-3.79\%} & [-6.62\% , -1.29\%] & \textcolor{dred}{2.17\%} & [1.04\% , 3.41\%] \\
Gavalas [benchmark] & 310.75 & 307.50 & 316.75 & \textcolor{dgreen}{-2.46\%} & [-4.32\% , -0.54\%] & \textcolor{dgreen}{-1.43\%} & [-3.07\% , 0.2\%] \\
\bottomrule 
\end{tabular}}
\caption{Average performance on the subset of instances (n=12 for Solomon, n=8 Cordeau and n=8 for Gavalas) of a global model trained on all instances and then fine-tuned during 50,000 epochs for each instance-region. Inference using beam search with a maximum number of 128 beams. The generated instances include n=64 generated instances for each of template benchmark instance-regions. The 95\% confidence intervals are computed using bootstrap (sampling instance-regions with replacement). For detailed performance on each individual benchmark instance and average performance for each instance-region see Tables \ref{table:full_results_bench_S}, \ref{table:full_results_gener_S}, \ref{table:full_results_bench_C}, \ref{table:full_results_gener_C},   \ref{table:full_results_bench_G} and  \ref{table:full_results_gener_G} in Supplementary materials.}
\label{table:sub_global}
\end{table}

\begin{table}[htbp]
\renewcommand{\arraystretch}{1.5} 
\centering
\resizebox{.55\width}{!}{\rowcolors{2}{black!5}{white}
\begin{tabular}{cccccccc}
\toprule 
\textbf{instance group} & \textbf{best-known} & \textbf{ILS} & \textbf{model score} & \textbf{gap to ILS} & \textbf{gap to ILS [95\%CI]} & \textbf{gap to bk} & \textbf{gap to bk [95\%CI]} \\
\midrule 
Solomon [generated] & - & 480.50 & 487.26 & \textcolor{dgreen}{-2.63\%} & [-3.23\% , -2.06\%] & - & - \\
Cordeau [generated] & - & 279.16 & 293.99 & \textcolor{dgreen}{-5.08\%} & [-5.74\% , -4.36\%] & - & - \\
Gavalas [generated] & - & 300.71 & 333.26 & \textcolor{dgreen}{-11.04\%} & [-12.09\% , -10.04\%] & - & - \\
Solomon [benchmark] & 624.59 & 607.12 & 607.55 & \textcolor{dgreen}{-0.35\%} & [-1.03\% , 0.29\%] & \textcolor{dred}{2.19\%} & [1.64\% , 2.78\%] \\
Cordeau [benchmark] & 498.50 & 458.60 & 482.75 & \textcolor{dgreen}{-4.77\%} & [-6.65\% , -2.95\%] & \textcolor{dred}{2.91\%} & [1.9\% , 4.01\%] \\
Gavalas [benchmark] & 307.64 & 298.88 & 311.15 & \textcolor{dgreen}{-4.52\%} & [-6.78\% , -2.54\%] & \textcolor{dgreen}{-0.82\%} & [-1.56\% , -0.09\%] \\
\bottomrule 
\end{tabular}}
\caption{Average performance on all instances of a global model trained on all instance-regions and then fine-tuned during 50,000 epochs for each instance-region. Inference using beam search with a maximum number of 128 beams. The generated instances include n=64 generated instances for each of template benchmark instance-regions. The 95\% confidence intervals are computed using bootstrap (sampling instance-regions with replacement). For detailed performance on each individual benchmark instance and average performance for each instance-region see Tables \ref{table:full_results_bench_S}, \ref{table:full_results_gener_S}, \ref{table:full_results_bench_C}, \ref{table:full_results_gener_C},   \ref{table:full_results_bench_G} and  \ref{table:full_results_gener_G} in Supplementary materials.}
\label{table:all_global}
\end{table}

\section{Discussion}
\label{sec:Discussion}

Here we show that a Pointer Network model trained using reinforcement learning is able to solve the Orienteering Problem with Time Windows. We approach the problem with the Tourist Trip Design Problem in mind, making sure that each model generalizes across different tourists who might visit the particular region(s) the model is trained on. We test our approach in both established and relatively new benchmark instances and show that it significantly outperforms the standard competitive heuristic ILS with inference times that are suitable for real time on-line applications. 

This work was inspired by previous application of PNs to solving np-hard combinatorial optimization problems such as the Travelling Salesman Problem and variations of the Vehicle Routing Problem, (\cite{Deudon2018, Nazari2018, Kool2019a}). We customized the PN architecture specifically for the OPTW. This customization happens mainly on the encoding of the points of interest, which is significantly different from previously proposed PNNs models. To start with, we use blocks of transformers (\cite{Deudon2018}) that being permutation invariant, unlike RNNs (\cite{Nazari2018}), are well suited for set encoding. Furthermore, we apply set encoding in a dynamic or iterative way. In particular, we use dynamical features that are updated at each iteration and we use one step lookahead to construct a graph representation of the admissible points of interest and use its adjacency matrix for self-attention masking. Importantly, we introduce a change in the transformer architecture: we add recursion by making the \textit{key} of the attention sub-block dependent on the previous iteration step. While these changes seem to improve the model (see Supplementary materials \ref{app:notweaks}), future work could offer a more systematic study on how these variations impact the model’s performance in this and other applications of the PNNs.

We have shown that we can get production-level performance and inference times. While ultimately the inference time is more important, the long training time can be inconvenient. Our main results are for a model trained for 500,000 epochs, which might take up to 72 hours. We show that training for $10\%$ of the epochs produces higher scores than the ILS algorithm and we have explored several ways of improving training speed and performance. In particular, we show that having a better generative model of tourists or a better sampling strategy improves training, and we also show, using different practical training scenarios, that using warmed-up models, transfer learning and fine-tuning works and speeds-up training.

The first application of PNNs for solving combinatorial optimization problems (\cite{Vinyals2015}) used supervised learning on data generated by algorithmic/heuristic approaches instead of reinforcement learning. This limits performance to how good the heuristics are. It would be interesting to see if such strategy could be used to warm-up a model for a reinforcement learning approach. 

A hybrid supervised and reinforcement learning approach could also benefit from access to real data. Our work shows that neural networks can be used in real world applications of the TTDP problem. The possibility of having access to real data, e.g., tours and tourist feedback, could lead to models that enhance route usability and could better address a tourist’s preferences. The formal optimal solution of the abstract optimization problem may not always be the best practical solution in a real-world setting. The ideal tour may depend on many other features not considered in the OPTW, which could be implicitly addressed in a machine learning approach that is capable of leveraging real data. 
The exploration strategy in a reinforcement learning approach can also have a big impact on performance and learning speed. While we use beam search for inference, we use stochastic sampling as an exploration strategy. We briefly explored using stochastic beam search but obtained worse results. It would be interesting to try different exploration strategies, like for instance a Monte Carlo tree search (\cite{Browne2012}). Another aspect of our exploration strategy is that each batch has samples from the same generated instance or tourist. It would be interesting to see if the inclusion of more tourists in the same batch (i.e. $k>1$ in \cite{Kool2019b}) can improve learning.

An advantage of a machine learning approach is that the output is probabilistic. This allows us, for instance, to retrieve ranked top-n solutions and use them in a broader route recommendation system. It also allows different inference strategies and the selection of the inference times according to the execution time requirements for the application at hand: we can make it faster by using almost instantaneous inference strategies like stochastic sampling or greedy inference, or by reducing the number of beams in beam search, or slower, but with better performance, by adding active search. This is an advantage over the several algorithmic approaches to the OPTW (\cite{OPbook}) that often have execution times that make it difficult to include them in real-time applications. 

Our approach is broad and flexible enough to be applied to different OPTW applications, like vehicle routing, transportation, scheduling, telecommunication, logistics and/or to other combinatorial optimization problems. The overall model architecture would remain the same. It would require the rewriting of admissibility conditions for node masking and lookahead search that are specific to the new optimization problem. It would also be necessary to choose features that make sense and are relevant for that specific problem. Furthermore, even for different OPTW applications the distribution from which training instances are sampled should reflect the specific application of interest. Altogether, the flexibility of the approach makes it a potentially useful tool for solving practical problems.

The source code can be found in GitHub \footnote{\url{https://github.com/mustelideos/optw_rl}}. 
 
 \section*{Acknowledgements}
 
The authors are grateful to Nishan Mann, Bahram Marami, João Gama and Hugo Penedones for their valuable suggestions. The first author is grateful to the Research Centre in Digital Services (CISeD), the Polytechnic of Viseu and FCT - Foundation for Science and Technology, I.P., within the scope of the project Refª UIDB/05583/2020 for their support.
 
\bibliographystyle{apalike}
\bibliography{bibfile.bib} 

\newpage

\appendix
 \section{Model Comparison: Recursion in the Transformer and Graph Masked Self-Attention}
\label{app:notweaks}
 
\begin{figure}[!htb]
\includegraphics[width=0.95\textwidth]{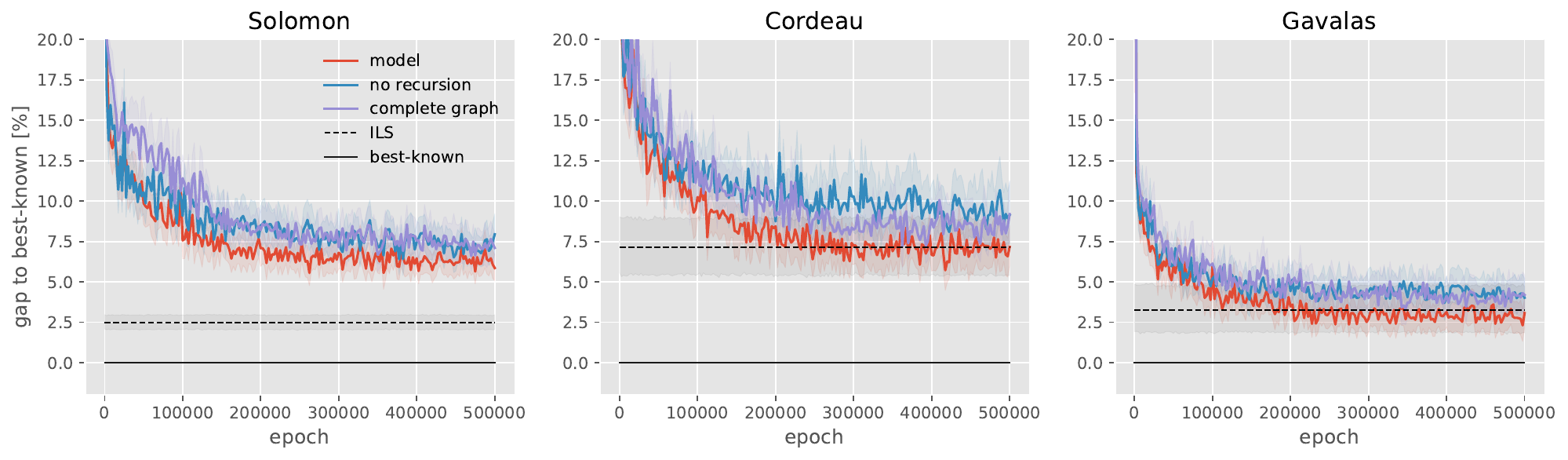}
\centering
\caption{Evolution of model performance (gap to best-known) during training using greedy inference, for the three models: proposed model (red), without recursion (blue), and without taking into account the look-ahead graph structure for masking (purple). The average model score gap and ILS-score gap to best-known are computed over all benchmark instances of the correspondent group (n=56 for Solomon, n=20 for Cordeau and n=33 for Gavalas). 
The shaded area represents 90\% confidence intervals (bootstrap benchmark instances).}
\label{fig:model_comparison_learning_benchmark}
\end{figure}

\begin{figure}[!htb]
\includegraphics[width=0.95\textwidth]{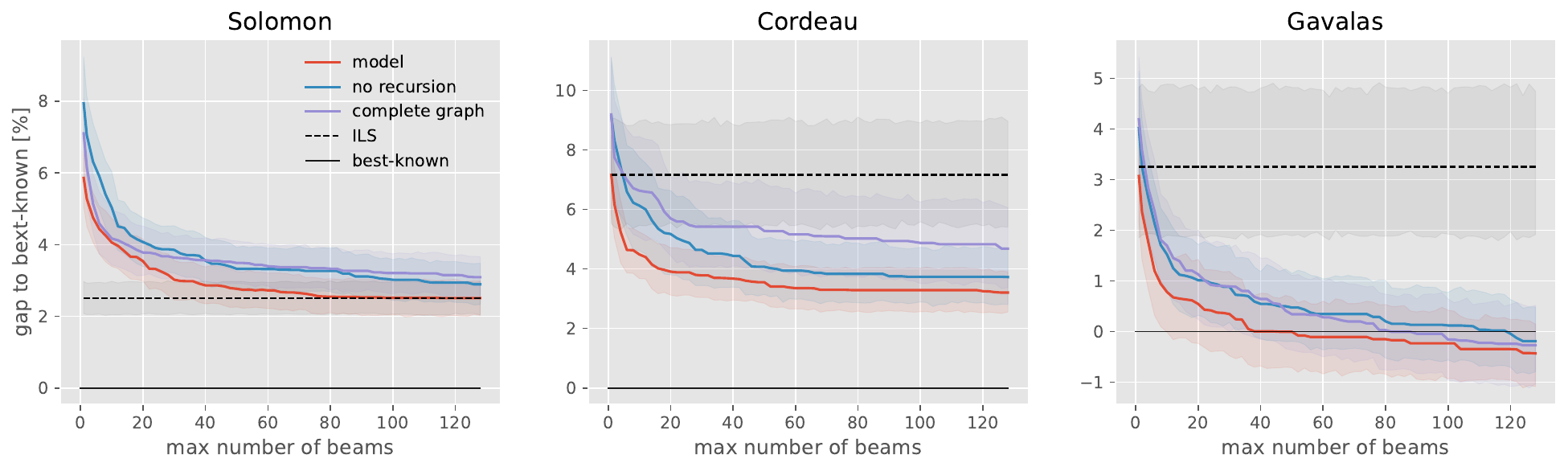}
\centering
\caption{Evolution of model performance (gap to  best-known) using beam search for inference with up to a maximum number of 128 beams, for the three models: proposed model (red), without recursion (blue), and without taking into account the look-ahead graph structure for masking (purple).
The average score gap and ILS-score gap to the best-known scores are computed over all benchmark instances of the correspondent group (n=56 for Solomon, n=20 for Cordeau and n=33 for Gavalas). 
The shaded area represents 90\% confidence intervals (bootstrap benchmark instances).}
\label{fig:model_comparison_beamsearch_benchmark}
\end{figure}

\begin{figure}[!htb]
\includegraphics[width=0.95\textwidth]{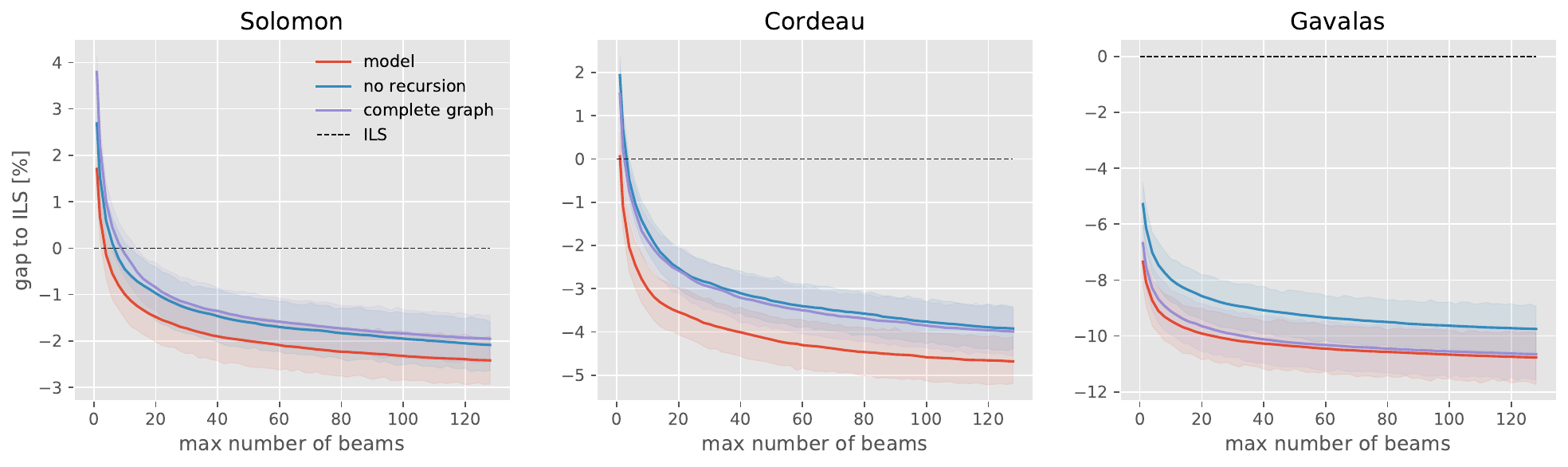}
\centering
\caption{Performance (gap to ILS) using beam search for inference with up to a maximum number of 128 beams, for the three models: proposed model (red), without recursion (blue), and without taking into account the look-ahead graph structure for masking (purple). The average score gap to the ILS score is computed over the mean score of all generated instances (n=64 generated instances per instance-region). The (very-)shaded area represents 90\% confidence intervals (bootstrap instance-regions).}
\label{fig:model_comparison_beamsearch_generated}
\vspace*{6in}
\end{figure}

We introduce some changes to the Pointer Neural Network used in previous applications to combinatorial optimization problems (see Section \ref{sec:review}). In this section we investigate the impact of two of those changes. Specifically, we compare our model to the exact same model but without recursion in the transformer (see Section \ref{sec:transformer_recursion}, and "no recursion" in Figures \ref{fig:model_comparison_learning_benchmark}, \ref{fig:model_comparison_beamsearch_benchmark} and \ref{fig:model_comparison_beamsearch_generated}) and to the exact same model but without the look-ahead graph masked self-attention (see Section \ref{sec:graph}, and "complete graph" in Figures \ref{fig:model_comparison_learning_benchmark}, \ref{fig:model_comparison_beamsearch_benchmark} and \ref{fig:model_comparison_beamsearch_generated}). For this model comparison we use the global model training scheme without fine-tuning (see Section \ref{subsec:experiments}), i.e., we train a model for each group of instances on all instances of each group at the same time: in each epoch, a tourist-instance is generated for/from a benchmark instance-region template sampled randomly from all instances of that group. We find that our model outperforms the models that lack each of those two changes. The difference can be observed during training (using greedy inference) (Fig.  \ref{fig:model_comparison_learning_benchmark}) and after training for beam search inference (Figs. \ref{fig:model_comparison_beamsearch_benchmark} and \ref{fig:model_comparison_beamsearch_generated}) 
on both the benchmark and generated instances.

\clearpage

 \section{Detailed Model Performance for All Instances}
\label{app:allresults}

\subsection{Model Trained from Scratch}
\label{app:modelresults}

Here we present the detailed individual results for the subset of 28 instances (n=12 for Solomon, n=8 for Cordeau and n=8 for Gavalas, see Section \ref{sec:methods}.  Methods for details). We evaluate performance on both benchmark (Table \ref{table:benchmark_results}) and generated (Table \ref{table:generated_results}) instances for the model(s) trained from scratch for 500,000 epochs and using beam search ("model") inference or active search followed by beam search ("model+as").
 
\subsubsection*{Benchmark Instances} 
\begin{table}[H]
\renewcommand{\arraystretch}{1.5} 
\centering
\resizebox{.5\width}{!}{\begin{tabular}{cccccccccc}
\toprule 
\textbf{group} & \textbf{instance} & \textbf{best-known (bk)} & \textbf{ILS} & \textbf{model} & \textbf{model gap to ILS} & \textbf{model gap to bk} & \textbf{model+as} & \textbf{model+as gap to ILS} & \textbf{model+as gap to bk} \\
\midrule 
\multirow{12}{*}[-0.4ex]{\rotatebox{90}{Solomon}} & r101 & 198 & 182 & 198 & \textcolor{dgreen}{-8.79\%} & \textcolor{black}{0.00\%} & 198 & \textcolor{dgreen}{-8.79\%} & \textcolor{black}{0.00\%} \\
 & \cellcolor{black!5} r102 & \cellcolor{black!5} 286 & \cellcolor{black!5} 286 & \cellcolor{black!5} 286 & \cellcolor{black!5} \textcolor{black}{0.00\%} & \cellcolor{black!5} \textcolor{black}{0.00\%} & \cellcolor{black!5} 286 & \cellcolor{black!5} \textcolor{black}{0.00\%} & \cellcolor{black!5} \textcolor{black}{0.00\%} \\
 & r201 & 797 & 788 & 793 & \textcolor{dgreen}{-0.63\%} & \textcolor{dred}{0.50\%} & 794 & \textcolor{dgreen}{-0.76\%} & \textcolor{dred}{0.38\%} \\
 & \cellcolor{black!5} r202 & \cellcolor{black!5} 930 & \cellcolor{black!5} 880 & \cellcolor{black!5} 886 & \cellcolor{black!5} \textcolor{dgreen}{-0.68\%} & \cellcolor{black!5} \textcolor{dred}{4.73\%} & \cellcolor{black!5} 890 & \cellcolor{black!5} \textcolor{dgreen}{-1.14\%} & \cellcolor{black!5} \textcolor{dred}{4.30\%} \\
 & rc101 & 219 & 219 & 219 & \textcolor{black}{0.00\%} & \textcolor{black}{0.00\%} & 219 & \textcolor{black}{0.00\%} & \textcolor{black}{0.00\%} \\
 & \cellcolor{black!5} rc102 & \cellcolor{black!5} 266 & \cellcolor{black!5} 259 & \cellcolor{black!5} 266 & \cellcolor{black!5} \textcolor{dgreen}{-2.70\%} & \cellcolor{black!5} \textcolor{black}{0.00\%} & \cellcolor{black!5} 266 & \cellcolor{black!5} \textcolor{dgreen}{-2.70\%} & \cellcolor{black!5} \textcolor{black}{0.00\%} \\
 & rc201 & 795 & 780 & 794 & \textcolor{dgreen}{-1.79\%} & \textcolor{dred}{0.13\%} & 794 & \textcolor{dgreen}{-1.79\%} & \textcolor{dred}{0.13\%} \\
 & \cellcolor{black!5} rc202 & \cellcolor{black!5} 936 & \cellcolor{black!5} 882 & \cellcolor{black!5} 934 & \cellcolor{black!5} \textcolor{dgreen}{-5.90\%} & \cellcolor{black!5} \textcolor{dred}{0.21\%} & \cellcolor{black!5} 935 & \cellcolor{black!5} \textcolor{dgreen}{-6.01\%} & \cellcolor{black!5} \textcolor{dred}{0.11\%} \\
 & c101 & 320 & 320& 320 & \textcolor{black}{0.00\%} & \textcolor{black}{0.00\%} & 320 & \textcolor{black}{0.00\%} & \textcolor{black}{0.00\%} \\
 & \cellcolor{black!5} c102 & \cellcolor{black!5} 360 & \cellcolor{black!5} 360 & \cellcolor{black!5} 360 & \cellcolor{black!5} \textcolor{black}{0.00\%} & \cellcolor{black!5} \textcolor{black}{0.00\%} & \cellcolor{black!5} 360 & \cellcolor{black!5} \textcolor{black}{0.00\%} & \cellcolor{black!5} \textcolor{black}{0.00\%} \\
 & c201 & 870 & 840 & 870 & \textcolor{dgreen}{-3.57\%} & \textcolor{black}{0.00\%} & 870 & \textcolor{dgreen}{-3.57\%} & \textcolor{black}{0.00\%} \\
 & \cellcolor{black!5} c202 & \cellcolor{black!5} 930 & \cellcolor{black!5} 910 & \cellcolor{black!5} 930 & \cellcolor{black!5} \textcolor{dgreen}{-2.20\%} & \cellcolor{black!5} \textcolor{black}{0.00\%} & \cellcolor{black!5} 920 & \cellcolor{black!5} \textcolor{dgreen}{-1.10\%} & \cellcolor{black!5} \textcolor{dred}{1.08\%} \\
\hline
\multirow{8}{*}[-0.4ex]{\rotatebox{90}{Cordeau}}& pr01 & 308 & 304 & 308 & \textcolor{dgreen}{-1.32\%} & \textcolor{black}{0.00\%} & 308 & \textcolor{dgreen}{-1.32\%} & \textcolor{black}{0.00\%} \\
 & \cellcolor{black!5} pr02 & \cellcolor{black!5} 404 & \cellcolor{black!5} 385 & \cellcolor{black!5} 402 & \cellcolor{black!5} \textcolor{dgreen}{-4.42\%} & \cellcolor{black!5} \textcolor{dred}{0.50\%} & \cellcolor{black!5} 401 & \cellcolor{black!5} \textcolor{dgreen}{-4.16\%} & \cellcolor{black!5} \textcolor{dred}{0.74\%} \\
 & pr03 & 394 & 384 & 375 & \textcolor{dred}{2.34\%} & \textcolor{dred}{4.82\%} & 384 & \textcolor{black}{0.00\%} & \textcolor{dred}{2.54\%} \\
 & \cellcolor{black!5} pr04 & \cellcolor{black!5} 489 & \cellcolor{black!5} 447 & \cellcolor{black!5} 489 & \cellcolor{black!5} \textcolor{dgreen}{-9.40\%} & \cellcolor{black!5} \textcolor{black}{0.00\%} & \cellcolor{black!5} 489 & \cellcolor{black!5} \textcolor{dgreen}{-9.40\%} & \cellcolor{black!5} \textcolor{black}{0.00\%} \\
 & pr11 & 353 & 330 & 351 & \textcolor{dgreen}{-6.36\%} & \textcolor{dred}{0.57\%} & 351 & \textcolor{dgreen}{-6.36\%} & \textcolor{dred}{0.57\%} \\
 & \cellcolor{black!5} pr12 & \cellcolor{black!5} 442 & \cellcolor{black!5} 431 & \cellcolor{black!5} 434 & \cellcolor{black!5} \textcolor{dgreen}{-0.70\%} & \cellcolor{black!5} \textcolor{dred}{1.81\%} & \cellcolor{black!5} 436 & \cellcolor{black!5} \textcolor{dgreen}{-1.16\%} & \cellcolor{black!5} \textcolor{dred}{1.36\%} \\
 & pr13 & 467 & 450 & 446 & \textcolor{dred}{0.89\%} & \textcolor{dred}{4.50\%} & 448 & \textcolor{dred}{0.44\%} & \textcolor{dred}{4.07\%} \\
 & \cellcolor{black!5} pr14 & \cellcolor{black!5} 567 & \cellcolor{black!5} 482 & \cellcolor{black!5} 541 & \cellcolor{black!5} \textcolor{dgreen}{-12.24\%} & \cellcolor{black!5} \textcolor{dred}{4.59\%} & \cellcolor{black!5} 545 & \cellcolor{black!5} \textcolor{dgreen}{-13.07\%} & \cellcolor{black!5} \textcolor{dred}{3.88\%} \\
\hline
\multirow{8}{*}[-0.4ex]{\rotatebox{90}{Gavalas}} & t101 & 387 & 387 & 399 & \textcolor{dgreen}{-3.10\%} & \textcolor{dgreen}{-3.10\%} & 399 & \textcolor{dgreen}{-3.10\%} & \textcolor{dgreen}{-3.10\%} \\
 & \cellcolor{black!5} t105 & \cellcolor{black!5} 433 & \cellcolor{black!5} 433 & \cellcolor{black!5} 427 & \cellcolor{black!5} \textcolor{dred}{1.39\%} & \cellcolor{black!5} \textcolor{dred}{1.39\%} & \cellcolor{black!5} 429 & \cellcolor{black!5} \textcolor{dred}{0.92\%} & \cellcolor{black!5} \textcolor{dred}{0.92\%} \\
 & t114 & 476 & 467 & 489 & \textcolor{dgreen}{-4.71\%} & \textcolor{dgreen}{-2.73\%} & 491 & \textcolor{dgreen}{-5.14\%} & \textcolor{dgreen}{-3.15\%} \\
 & \cellcolor{black!5} t117 & \cellcolor{black!5} 462 & \cellcolor{black!5} 452 & \cellcolor{black!5} 470 & \cellcolor{black!5} \textcolor{dgreen}{-3.98\%} & \cellcolor{black!5} \textcolor{dgreen}{-1.73\%} & \cellcolor{black!5} 471 & \cellcolor{black!5} \textcolor{dgreen}{-4.20\%} & \cellcolor{black!5} \textcolor{dgreen}{-1.95\%} \\
 & t201 & 185 & 183 & 183 & \textcolor{black}{0.00\%} & \textcolor{dred}{1.08\%} & 191 & \textcolor{dgreen}{-4.37\%} & \textcolor{dgreen}{-3.24\%} \\
 & \cellcolor{black!5} t202 & \cellcolor{black!5} 193 & \cellcolor{black!5} 193 & \cellcolor{black!5} 188 & \cellcolor{black!5} \textcolor{dred}{2.59\%} & \cellcolor{black!5} \textcolor{dred}{2.59\%} & \cellcolor{black!5} 193 & \cellcolor{black!5} \textcolor{black}{0.00\%} & \cellcolor{black!5} \textcolor{black}{0.00\%} \\
 & t203 & 179 & 174 & 181 & \textcolor{dgreen}{-4.02\%} & \textcolor{dgreen}{-1.12\%} & 181 & \textcolor{dgreen}{-4.02\%} & \textcolor{dgreen}{-1.12\%} \\
 & \cellcolor{black!5} t204 & \cellcolor{black!5} 171 & \cellcolor{black!5} 171 & \cellcolor{black!5} 174 & \cellcolor{black!5} \textcolor{dgreen}{-1.75\%} & \cellcolor{black!5} \textcolor{dgreen}{-1.75\%} & \cellcolor{black!5} 174 & \cellcolor{black!5} \textcolor{dgreen}{-1.75\%} & \cellcolor{black!5} \textcolor{dgreen}{-1.75\%} \\
\bottomrule 
\end{tabular}}
\caption{Performance (score, gap to ILS and gap to best-known) for models trained from scratch for 500,000 epochs using beam-search for inference ("model") or 128 epochs of active search before beam-search ("model+as"), for each of the benchmark instances in the subset of instances (12 Solomon, 8 Cordeau and 8 Gavalas).}
\label{table:benchmark_results}
\vspace*{1in}
\end{table}

\subsubsection*{Generated Instances} 
\begin{table}[H]
\renewcommand{\arraystretch}{1.5} 
\centering
\resizebox{.5\width}{!}{\begin{tabular}{ccccccc}
\toprule 
\textbf{group} & \textbf{instance [generated]} & \textbf{ILS} & \textbf{model} & \textbf{gap to ILS} & \textbf{model+as} & \textbf{model+as gap to ILS} \\
\midrule 
\multirow{12}{*}[-0.4ex]{\rotatebox{90}{Solomon}} & r101 & 108.34 & 115.45 & \textcolor{dgreen}{-6.56\%} & 115.45 & \textcolor{dgreen}{-6.56\%} \\
 & \cellcolor{black!5} r102 & \cellcolor{black!5} 155.72 & \cellcolor{black!5} 166.19 & \cellcolor{black!5} \textcolor{dgreen}{-6.72\%} & \cellcolor{black!5} 165.97 & \cellcolor{black!5} \textcolor{dgreen}{-6.58\%} \\
 & r201 & 619.38 & 627.86 & \textcolor{dgreen}{-1.37\%} & 630.36 & \textcolor{dgreen}{-1.77\%} \\
 & \cellcolor{black!5} r202 & \cellcolor{black!5} 790.61 & \cellcolor{black!5} 800.88 & \cellcolor{black!5} \textcolor{dgreen}{-1.30\%} & \cellcolor{black!5} 805.11 & \cellcolor{black!5} \textcolor{dgreen}{-1.83\%} \\
 & rc101 & 152.08 & 159.86 & \textcolor{dgreen}{-5.12\%} & 159.98 & \textcolor{dgreen}{-5.20\%} \\
 & \cellcolor{black!5} rc102 & \cellcolor{black!5} 179.44 & \cellcolor{black!5} 185.72 & \cellcolor{black!5} \textcolor{dgreen}{-3.50\%} & \cellcolor{black!5} 186.44 & \cellcolor{black!5} \textcolor{dgreen}{-3.90\%} \\
 & rc201 & 517.08 & 527.36 & \textcolor{dgreen}{-1.99\%} & 529.02 & \textcolor{dgreen}{-2.31\%} \\
 & \cellcolor{black!5} rc202 & \cellcolor{black!5} 651.62 & \cellcolor{black!5} 661.44 & \cellcolor{black!5} \textcolor{dgreen}{-1.51\%} & \cellcolor{black!5} 663.59 & \cellcolor{black!5} \textcolor{dgreen}{-1.84\%} \\
 & c101 & 255.92 & 265.61 & \textcolor{dgreen}{-3.79\%} & 265.73 & \textcolor{dgreen}{-3.83\%} \\
 & \cellcolor{black!5} c102 & \cellcolor{black!5} 290.80 & \cellcolor{black!5} 298.00 & \cellcolor{black!5} \textcolor{dgreen}{-2.48\%} & \cellcolor{black!5} 298.55 & \cellcolor{black!5} \textcolor{dgreen}{-2.67\%} \\
 & c201 & 687.45 & 693.98 & \textcolor{dgreen}{-0.95\%} & 694.03 & \textcolor{dgreen}{-0.96\%} \\
 & \cellcolor{black!5} c202 & \cellcolor{black!5} 751.44 & \cellcolor{black!5} 762.25 & \cellcolor{black!5} \textcolor{dgreen}{-1.44\%} & \cellcolor{black!5} 762.92 & \cellcolor{black!5} \textcolor{dgreen}{-1.53\%} \\
\hline
\multirow{8}{*}[-0.4ex]{\rotatebox{90}{Cordeau}} & pr01 & 191.72 & 199.64 & \textcolor{dgreen}{-4.13\%} & 199.48 & \textcolor{dgreen}{-4.05\%} \\
 & \cellcolor{black!5} pr02 & \cellcolor{black!5} 204.42 & \cellcolor{black!5} 213.53 & \cellcolor{black!5} \textcolor{dgreen}{-4.46\%} & \cellcolor{black!5} 214.00 & \cellcolor{black!5} \textcolor{dgreen}{-4.69\%} \\
 & pr03 & 246.08 & 263.34 & \textcolor{dgreen}{-7.02\%} & 263.41 & \textcolor{dgreen}{-7.04\%} \\
 & \cellcolor{black!5} pr04 & \cellcolor{black!5} 304.45 & \cellcolor{black!5} 323.97 & \cellcolor{black!5} \textcolor{dgreen}{-6.41\%} & \cellcolor{black!5} 324.77 & \cellcolor{black!5} \textcolor{dgreen}{-6.67\%} \\
 & pr11 & 208.31 & 216.05 & \textcolor{dgreen}{-3.71\%} & 216.31 & \textcolor{dgreen}{-3.84\%} \\
 & \cellcolor{black!5} pr12 & \cellcolor{black!5} 233.09 & \cellcolor{black!5} 239.44 & \cellcolor{black!5} \textcolor{dgreen}{-2.72\%} & \cellcolor{black!5} 239.94 & \cellcolor{black!5} \textcolor{dgreen}{-2.94\%} \\
 & pr13 & 294.72 & 304.38 & \textcolor{dgreen}{-3.28\%} & 306.45 & \textcolor{dgreen}{-3.98\%} \\
 & \cellcolor{black!5} pr14 & \cellcolor{black!5} 346.69 & \cellcolor{black!5} 372.55 & \cellcolor{black!5} \textcolor{dgreen}{-7.46\%} & \cellcolor{black!5} 374.23 & \cellcolor{black!5} \textcolor{dgreen}{-7.95\%} \\
\hline
\multirow{8}{*}[-0.4ex]{\rotatebox{90}{Gavalas}} & t101 & 320.44 & 339.16 & \textcolor{dgreen}{-5.84\%} & 339.78 & \textcolor{dgreen}{-6.04\%} \\
 & \cellcolor{black!5} t105 & \cellcolor{black!5} 363.44 & \cellcolor{black!5} 393.23 & \cellcolor{black!5} \textcolor{dgreen}{-8.20\%} & \cellcolor{black!5} 394.58 & \cellcolor{black!5} \textcolor{dgreen}{-8.57\%} \\
 & t114 & 328.36 & 358.55 & \textcolor{dgreen}{-9.19\%} & 360.22 & \textcolor{dgreen}{-9.70\%} \\
 & \cellcolor{black!5} t117 & \cellcolor{black!5} 401.27 & \cellcolor{black!5} 457.23 & \cellcolor{black!5} \textcolor{dgreen}{-13.95\%} & \cellcolor{black!5} 458.14 & \cellcolor{black!5} \textcolor{dgreen}{-14.17\%} \\
 & t201 & 235.88 & 266.36 & \textcolor{dgreen}{-12.92\%} & 266.28 & \textcolor{dgreen}{-12.89\%} \\
 & \cellcolor{black!5} t202 & \cellcolor{black!5} 252.59 & \cellcolor{black!5} 302.53 & \cellcolor{black!5} \textcolor{dgreen}{-19.77\%} & \cellcolor{black!5} 302.92 & \cellcolor{black!5} \textcolor{dgreen}{-19.92\%} \\
 & t203 & 250.39 & 273.94 & \textcolor{dgreen}{-9.40\%} & 274.28 & \textcolor{dgreen}{-9.54\%} \\
 & \cellcolor{black!5} t204 & \cellcolor{black!5} 247.53 & \cellcolor{black!5} 279.39 & \cellcolor{black!5} \textcolor{dgreen}{-12.87\%} & \cellcolor{black!5} 279.78 & \cellcolor{black!5} \textcolor{dgreen}{-13.03\%} \\
\bottomrule 
\end{tabular}}
\caption{Average performance (score and gap to ILS) for models trained from scratch for 500,000 epochs using beam-search for inference ("model") or 128 epochs of active search before beam-search ("model+as"), for each of the generated instance-regions in the subset of instances (12 Solomon, 8 Cordeau and 8 Gavalas). The values are averages across the n=64 generated instances for each of template benchmark instance-regions.}
\label{table:generated_results}
\vspace*{1in}
\end{table}

\subsection{Fine-tuned Global Model}
\label{app:globalresults}

We want to present results for all the instance-regions of each group of benchmark instances without having to go through the time-consuming process of training from scratch each and every one of them. In order to achieve that goal, we train a global model for each instance group ,i.e., a model trained simultaneously on all the instance-regions of that group, and then fine-tune it for $10\%$ (50,000) of the epochs on each instance-region (see "Fine-tune a global model for every instance" in Section \ref{subseb:global}).

\subsubsection*{Solomon [Benchmark]}

\begin{table}[H]
\renewcommand{\arraystretch}{1.5} 
\centering
\resizebox{.5\width}{!}{\rowcolors{2}{black!5}{white}
\begin{tabular}{ccccccccccccc}
\toprule 
\textbf{instance} & \textbf{best-known} & \textbf{ILS} & \textbf{global+ft} & \textbf{gap to ILS} & \textbf{gap to best-known} & \textbf{} & \textbf{instance} & \textbf{best-known} & \textbf{ILS} & \textbf{global+ft} & \textbf{gap to ILS} & \textbf{gap to best-known} \\
\midrule 
c101 & 320 & 320 & 320 & \textcolor{black}{0.00\%} & \textcolor{black}{0.00\%} &  & c201 & 870 & 840 & 870 & \textcolor{dgreen}{-3.57\%} & \textcolor{black}{0.00\%} \\
c102 & 360 & 360 & 360 & \textcolor{black}{0.00\%} & \textcolor{black}{0.00\%} &  & c202 & 930 & 910 & 930 & \textcolor{dgreen}{-2.20\%} & \textcolor{black}{0.00\%} \\
c103 & 400 & 390 & 390 & \textcolor{black}{0.00\%} & \textcolor{dred}{2.50\%} &  & c203 & 960 & 940 & 960 & \textcolor{dgreen}{-2.13\%} & \textcolor{black}{0.00\%} \\
c104 & 420 & 400 & 410 & \textcolor{dgreen}{-2.50\%} & \textcolor{dred}{2.38\%} &  & c204 & 980 & 950 & 970 & \textcolor{dgreen}{-2.11\%} & \textcolor{dred}{1.02\%} \\
c105 & 340 & 340 & 340 & \textcolor{black}{0.00\%} & \textcolor{black}{0.00\%} &  & c205 & 910 & 900 & 890 & \textcolor{dred}{1.11\%} & \textcolor{dred}{2.20\%} \\
c106 & 340 & 340 & 340 & \textcolor{black}{0.00\%} & \textcolor{black}{0.00\%} &  & c206 & 930 & 910 & 920 & \textcolor{dgreen}{-1.10\%} & \textcolor{dred}{1.08\%} \\
c107 & 370 & 360 & 370 & \textcolor{dgreen}{-2.78\%} & \textcolor{black}{0.00\%} &  & c207 & 930 & 910 & 910 & \textcolor{black}{0.00\%} & \textcolor{dred}{2.15\%} \\
c108 & 370 & 370 & 370 & \textcolor{black}{0.00\%} & \textcolor{black}{0.00\%} & & c208 & 950 & 930 & 940 & \textcolor{dgreen}{-1.08\%} & \textcolor{dred}{1.05\%} \\
c109 & 380 & 380 & 380 & \textcolor{black}{0.00\%} & \textcolor{black}{0.00\%} &  & - & - & - & - & - & - \\
r101 & 198 & 182 & 198 & \textcolor{dgreen}{-8.79\%} & \textcolor{black}{0.00\%} &  & r201 & 797 & 788 & 778 & \textcolor{dred}{1.27\%} & \textcolor{dred}{2.38\%} \\
r102 & 286 & 286 & 286 & \textcolor{black}{0.00\%} & \textcolor{black}{0.00\%} &  & r202 & 930 & 880 & 879 & \textcolor{dred}{0.11\%} & \textcolor{dred}{5.48\%} \\
r103 & 293 & 286 & 291 & \textcolor{dgreen}{-1.75\%} & \textcolor{dred}{0.68\%} &  & r203 & 1028 & 980 & 967 & \textcolor{dred}{1.33\%} & \textcolor{dred}{5.93\%} \\
r104 & 303 & 297 & 299 & \textcolor{dgreen}{-0.67\%} & \textcolor{dred}{1.32\%} &  & r204 & 1093 & 1073 & 1035 & \textcolor{dred}{3.54\%} & \textcolor{dred}{5.31\%} \\
r105 & 247 & 247 & 247 & \textcolor{black}{0.00\%} & \textcolor{black}{0.00\%} &  & r205 & 953 & 931 & 927 & \textcolor{dred}{0.43\%} & \textcolor{dred}{2.73\%} \\
r106 & 293 & 293 & 288 & \textcolor{dred}{1.71\%} & \textcolor{dred}{1.71\%} &  & r206 & 1032 & 996 & 949 & \textcolor{dred}{4.72\%} & \textcolor{dred}{8.04\%} \\
r107 & 299 & 288 & 295 & \textcolor{dgreen}{-2.43\%} & \textcolor{dred}{1.34\%} &  & r207 & 1077 & 1038 & 1017 & \textcolor{dred}{2.02\%} & \textcolor{dred}{5.57\%} \\
r108 & 308 & 297 & 303 & \textcolor{dgreen}{-2.02\%} & \textcolor{dred}{1.62\%} &  & r208 & 1117 & 1069 & 1062 & \textcolor{dred}{0.65\%} & \textcolor{dred}{4.92\%} \\
r109 & 277 & 276 & 277 & \textcolor{dgreen}{-0.36\%} & \textcolor{black}{0.00\%} &  & r209 & 961 & 926 & 913 & \textcolor{dred}{1.40\%} & \textcolor{dred}{4.99\%} \\
r110 & 284 & 281 & 283 & \textcolor{dgreen}{-0.71\%} & \textcolor{dred}{0.35\%} &  & r210 & 1000 & 958 & 950 & \textcolor{dred}{0.84\%} & \textcolor{dred}{5.00\%} \\
r111 & 297 & 295 & 294 & \textcolor{dred}{0.34\%} & \textcolor{dred}{1.01\%} &  & r211 & 1051 & 1023 & 1027 & \textcolor{dgreen}{-0.39\%} & \textcolor{dred}{2.28\%} \\
r112 & 298 & 295 & 291 & \textcolor{dred}{1.36\%} & \textcolor{dred}{2.35\%} &  & - & - & - & - & - & - \\
rc101 & 219 & 219 & 219 & \textcolor{black}{0.00\%} & \textcolor{black}{0.00\%} &  & rc201 & 795 & 780 & 788 & \textcolor{dgreen}{-1.03\%} & \textcolor{dred}{0.88\%} \\
rc102 & 266 & 259 & 259 & \textcolor{black}{0.00\%} & \textcolor{dred}{2.63\%} &  & rc202 & 936 & 882 & 906 & \textcolor{dgreen}{-2.72\%} & \textcolor{dred}{3.21\%} \\
rc103 & 266 & 265 & 263 & \textcolor{dred}{0.75\%} & \textcolor{dred}{1.13\%} &  & rc203 & 1003 & 960 & 967 & \textcolor{dgreen}{-0.73\%} & \textcolor{dred}{3.59\%} \\
rc104 & 301 & 297 & 277 & \textcolor{dred}{6.73\%} & \textcolor{dred}{7.97\%} &  & rc204 & 1140 & 1117 & 1086 & \textcolor{dred}{2.78\%} & \textcolor{dred}{4.74\%} \\
rc105 & 244 & 221 & 241 & \textcolor{dgreen}{-9.05\%} & \textcolor{dred}{1.23\%} &  & rc205 & 859 & 840 & 847 & \textcolor{dgreen}{-0.83\%} & \textcolor{dred}{1.40\%} \\
rc106 & 252 & 239 & 245 & \textcolor{dgreen}{-2.51\%} & \textcolor{dred}{2.78\%} &  & rc206 & 899 & 860 & 875 & \textcolor{dgreen}{-1.74\%} & \textcolor{dred}{2.67\%} \\
rc107 & 277 & 274 & 274 & \textcolor{black}{0.00\%} & \textcolor{dred}{1.08\%} &  & rc207 & 983 & 926 & 936 & \textcolor{dgreen}{-1.08\%} & \textcolor{dred}{4.78\%} \\
rc108 & 298 & 288 & 277 & \textcolor{dred}{3.82\%} & \textcolor{dred}{7.05\%} &  & rc208 & 1057 & 1037 & 1037 & \textcolor{black}{0.00\%} & \textcolor{dred}{1.89\%} \\
\bottomrule 
\end{tabular}}
\caption{Performance (score, gap to ILS and gap to best-known) on Solomon benchmark instances of a global model trained on all instance-regions of this group and then fine-tuned during 50,000 epochs for each instance-region ("global+ft"). Inference uses beam search with a maximum of 128 beams.}
\label{table:full_results_bench_S}
\vspace*{3in}
\end{table}

\subsubsection*{Solomon [Generated]}
\begin{table}[H]
\renewcommand{\arraystretch}{1.5} 
\centering
\resizebox{.5\width}{!}{\rowcolors{2}{black!5}{white}
\begin{tabular}{ccccccccc}
\toprule 
\textbf{instance [generated]} & \textbf{ILS} & \textbf{global+ft} & \textbf{global+ft gap to ILS} &  & \textbf{instance [generated]} & \textbf{ILS} & \textbf{global+ft} & \textbf{global+ft gap to ILS} \\
\midrule 
c101 & 255.92 & 266.19 & \textcolor{dgreen}{-4.01\%} &  & c201 & 687.45 & 693.94 & \textcolor{dgreen}{-0.94\%} \\
c102 & 290.80 & 299.03 & \textcolor{dgreen}{-2.83\%} &  & c202 & 751.44 & 760.77 & \textcolor{dgreen}{-1.24\%} \\
c103 & 303.86 & 311.31 & \textcolor{dgreen}{-2.45\%} &  & c203 & 790.70 & 798.83 & \textcolor{dgreen}{-1.03\%} \\
c104 & 312.19 & 320.19 & \textcolor{dgreen}{-2.56\%} &  & c204 & 826.61 & 828.31 & \textcolor{dgreen}{-0.21\%} \\
c105 & 273.08 & 282.61 & \textcolor{dgreen}{-3.49\%} &  & c205 & 725.75 & 731.25 & \textcolor{dgreen}{-0.76\%} \\
c106 & 279.09 & 286.17 & \textcolor{dgreen}{-2.54\%} &  & c206 & 747.22 & 751.27 & \textcolor{dgreen}{-0.54\%} \\
c107 & 284.92 & 291.75 & \textcolor{dgreen}{-2.40\%} &  & c207 & 760.66 & 764.17 & \textcolor{dgreen}{-0.46\%} \\
c108 & 288.83 & 296.48 & \textcolor{dgreen}{-2.65\%} &  & c208 & 763.56 & 769.16 & \textcolor{dgreen}{-0.73\%} \\
c109 & 299.25 & 306.02 & \textcolor{dgreen}{-2.26\%} &  & - & - & - & - \\
r101 & 108.34 & 115.45 & \textcolor{dgreen}{-6.56\%} &  & r201 & 619.38 & 625.62 & \textcolor{dgreen}{-1.01\%} \\
r102 & 155.72 & 166.20 & \textcolor{dgreen}{-6.73\%} &  & r202 & 790.61 & 796.06 & \textcolor{dgreen}{-0.69\%} \\
r103 & 178.41 & 188.78 & \textcolor{dgreen}{-5.82\%} &  & r203 & 887.61 & 888.34 & \textcolor{dgreen}{-0.08\%} \\
r104 & 193.25 & 204.77 & \textcolor{dgreen}{-5.96\%} &  & r204 & 959.69 & 968.80 & \textcolor{dgreen}{-0.95\%} \\
r105 & 140.53 & 147.61 & \textcolor{dgreen}{-5.04\%} &  & r205 & 759.66 & 765.20 & \textcolor{dgreen}{-0.73\%} \\
r106 & 170.23 & 178.41 & \textcolor{dgreen}{-4.80\%} &  & r206 & 870.16 & 876.39 & \textcolor{dgreen}{-0.72\%} \\
r107 & 185.20 & 195.53 & \textcolor{dgreen}{-5.58\%} &  & r207 & 927.34 & 929.77 & \textcolor{dgreen}{-0.26\%} \\
r108 & 195.86 & 207.61 & \textcolor{dgreen}{-6.00\%} &  & r208 & 982.83 & 985.75 & \textcolor{dgreen}{-0.30\%} \\
r109 & 164.33 & 170.86 & \textcolor{dgreen}{-3.97\%} &  & r209 & 826.05 & 826.77 & \textcolor{dgreen}{-0.09\%} \\
r110 & 175.50 & 184.83 & \textcolor{dgreen}{-5.32\%} &  & r210 & 852.52 & 856.11 & \textcolor{dgreen}{-0.42\%} \\
r111 & 177.42 & 188.92 & \textcolor{dgreen}{-6.48\%} &  & r211 & 894.78 & 898.62 & \textcolor{dgreen}{-0.43\%} \\
r112 & 188.23 & 199.56 & \textcolor{dgreen}{-6.02\%} &  & - & - & - & - \\
rc101 & 152.08 & 160.11 & \textcolor{dgreen}{-5.28\%} &  & rc201 & 517.08 & 525.31 & \textcolor{dgreen}{-1.59\%} \\
rc102 & 179.44 & 185.61 & \textcolor{dgreen}{-3.44\%} &  & rc202 & 651.62 & 656.78 & \textcolor{dgreen}{-0.79\%} \\
rc103 & 194.81 & 203.81 & \textcolor{dgreen}{-4.62\%} &  & rc203 & 747.47 & 743.02 & \textcolor{dred}{0.60\%} \\
rc104 & 207.03 & 219.67 & \textcolor{dgreen}{-6.11\%} &  & rc204 & 825.05 & 828.11 & \textcolor{dgreen}{-0.37\%} \\
rc105 & 168.97 & 177.41 & \textcolor{dgreen}{-4.99\%} &  & rc205 & 599.28 & 600.44 & \textcolor{dgreen}{-0.19\%} \\
rc106 & 174.31 & 182.31 & \textcolor{dgreen}{-4.59\%} &  & rc206 & 620.97 & 625.72 & \textcolor{dgreen}{-0.76\%} \\
rc107 & 189.89 & 198.09 & \textcolor{dgreen}{-4.32\%} &  & rc207 & 688.55 & 694.02 & \textcolor{dgreen}{-0.79\%} \\
rc108 & 198.75 & 205.44 & \textcolor{dgreen}{-3.36\%} &  & rc208 & 747.48 & 757.52 & \textcolor{dgreen}{-1.34\%} \\
\bottomrule 
\end{tabular}}
\caption{Average performance (score and gap to ILS) on Solomon generated instances of a global model trained on all instance-regions of this group and then fine-tuned during 50,000 epochs for each instance-region ("global+ft"). Inference uses beam search with a maximum of 128 beams. The values are averages across the n=64 generated instances.}
\label{table:full_results_gener_S}
\vspace*{3in}
\end{table}

\subsubsection*{Cordeau [Benchmark]}
\begin{table}[H]
\renewcommand{\arraystretch}{1.5} 
\centering
\resizebox{.5\width}{!}{\rowcolors{2}{black!5}{white}
\begin{tabular}{ccccccccccccc}
\toprule 
\textbf{instance} & \textbf{best-known} & \textbf{ILS} & \textbf{global+ft} & \textbf{gap to ILS} & \textbf{gap to best-known} & \textbf{} & \textbf{instance} & \textbf{best-known} & \textbf{ILS} & \textbf{global+ft} & \textbf{gap to ILS} & \textbf{gap to best-known} \\
\midrule 
pr01 & 308 & 304 & 308 & \textcolor{dgreen}{-1.32\%} & \textcolor{black}{0.00\%} &  & pr11 & 353 & 330 & 351 & \textcolor{dgreen}{-6.36\%} & \textcolor{dred}{0.57\%} \\
pr02 & 404 & 385 & 401 & \textcolor{dgreen}{-4.16\%} & \textcolor{dred}{0.74\%} &  & pr12 & 442 & 431 & 436 & \textcolor{dgreen}{-1.16\%} & \textcolor{dred}{1.36\%} \\
pr03 & 394 & 384 & 380 & \textcolor{dred}{1.04\%} & \textcolor{dred}{3.55\%} &  & pr13 & 467 & 450 & 453 & \textcolor{dgreen}{-0.67\%} & \textcolor{dred}{3.00\%} \\
pr04 & 489 & 447 & 476 & \textcolor{dgreen}{-6.49\%} & \textcolor{dred}{2.66\%} &  & pr14 & 567 & 482 & 536 & \textcolor{dgreen}{-11.20\%} & \textcolor{dred}{5.47\%} \\
pr05 & 595 & 576 & 591 & \textcolor{dgreen}{-2.60\%} & \textcolor{dred}{0.67\%} &  & pr15 & 708 & 638 & 699 & \textcolor{dgreen}{-9.56\%} & \textcolor{dred}{1.27\%} \\
pr06 & 591 & 538 & 567 & \textcolor{dgreen}{-5.39\%} & \textcolor{dred}{4.06\%} &  & pr16 & 674 & 559 & 609 & \textcolor{dgreen}{-8.94\%} & \textcolor{dred}{9.64\%} \\
pr07 & 298 & 291 & 293 & \textcolor{dgreen}{-0.69\%} & \textcolor{dred}{1.68\%} &  & pr17 & 362 & 346 & 349 & \textcolor{dgreen}{-0.87\%} & \textcolor{dred}{3.59\%} \\
pr08 & 463 & 463 & 461 & \textcolor{dred}{0.43\%} & \textcolor{dred}{0.43\%} &  & pr18 & 539 & 479 & 532 & \textcolor{dgreen}{-11.06\%} & \textcolor{dred}{1.30\%} \\
pr09 & 493 & 461 & 466 & \textcolor{dgreen}{-1.08\%} & \textcolor{dred}{5.48\%} &  & pr19 & 562 & 499 & 522 & \textcolor{dgreen}{-4.61\%} & \textcolor{dred}{7.12\%} \\
pr10 & 594 & 539 & 580 & \textcolor{dgreen}{-7.61\%} & \textcolor{dred}{2.36\%} &  & pr20 & 667 & 570 & 645 & \textcolor{dgreen}{-13.16\%} & \textcolor{dred}{3.30\%} \\
\bottomrule 
\end{tabular}}
\caption{Performance (score, gap to ILS and gap to best-known) on Cordeau instances of a global model trained on all instance-regions of this group and then fine-tuned during 50,000 epochs for each instance-region ("global+ft"). Inference uses beam search with a maximum of 128 beams.}
\label{table:full_results_bench_C}
\vspace*{1in}
\end{table}

\subsubsection*{Cordeau [Generated]}
\begin{table}[H]
\renewcommand{\arraystretch}{1.5} 
\centering
\resizebox{.5\width}{!}{\rowcolors{2}{black!5}{white}
\begin{tabular}{ccccccccc}
\toprule 
\textbf{instance [generated]} & \textbf{ILS} & \textbf{global+ft} & \textbf{global+ft gap to ILS} &   & \textbf{instance [generated]} & \textbf{ILS} & \textbf{global+ft} & \textbf{global+ft gap to ILS} \\
\midrule 
pr01 & 191.72 & 199.77 & \textcolor{dgreen}{-4.20\%} &  & pr11 & 208.31 & 215.72 & \textcolor{dgreen}{-3.56\%} \\
pr02 & 204.42 & 213.78 & \textcolor{dgreen}{-4.58\%} &  & pr12 & 233.09 & 238.17 & \textcolor{dgreen}{-2.18\%} \\
pr03 & 246.08 & 261.67 & \textcolor{dgreen}{-6.34\%} &  & pr13 & 294.72 & 304.12 & \textcolor{dgreen}{-3.19\%} \\
pr04 & 304.45 & 321.86 & \textcolor{dgreen}{-5.72\%} &  & pr14 & 346.69 & 369.67 & \textcolor{dgreen}{-6.63\%} \\
pr05 & 291.12 & 310.86 & \textcolor{dgreen}{-6.78\%} &  & pr15 & 332.14 & 354.41 & \textcolor{dgreen}{-6.70\%} \\
pr06 & 325.52 & 342.89 & \textcolor{dgreen}{-5.34\%} &  & pr16 & 363.42 & 391.25 & \textcolor{dgreen}{-7.66\%} \\
pr07 & 174.28 & 181.53 & \textcolor{dgreen}{-4.16\%} &  & pr17 & 208.69 & 212.42 & \textcolor{dgreen}{-1.79\%} \\
pr08 & 246.45 & 256.48 & \textcolor{dgreen}{-4.07\%} &  & pr18 & 274.47 & 286.84 & \textcolor{dgreen}{-4.51\%} \\
pr09 & 289.58 & 307.12 & \textcolor{dgreen}{-6.06\%} &  & pr19 & 325.72 & 345.70 & \textcolor{dgreen}{-6.14\%} \\
pr10 & 341.28 & 361.50 & \textcolor{dgreen}{-5.92\%} &  & pr20 & 381.05 & 403.94 & \textcolor{dgreen}{-6.01\%} \\
\bottomrule 
\end{tabular}}
\caption{Average performance (score and gap to ILS) on Cordeau generated instances of a global model trained on all instance-regions of this group and then fine-tuned during 50,000 epochs for each instance-region ("global+ft"). Inference uses beam search with a maximum of 128 beams. The values are averages across the n=64 generated instances.}
\label{table:full_results_gener_C}
\vspace*{1in}
\end{table}

\subsubsection*{Gavalas [Benchmark]}
\begin{table}[H]
\renewcommand{\arraystretch}{1.5} 
\centering
\resizebox{.5\width}{!}{\rowcolors{2}{black!5}{white}
\begin{tabular}{ccccccccccccc}
\toprule 
\textbf{instance} & \textbf{best-known} & \textbf{ILS} & \textbf{global+ft} & \textbf{gap to ILS} & \textbf{gap to best-known} & \textbf{ } & \textbf{instance} & \textbf{best-known} & \textbf{ILS} & \textbf{global+ft} & \textbf{gap to ILS} & \textbf{gap to best-known} \\
\midrule 
t101 & 387 & 387 & 402 & \textcolor{dgreen}{-3.88\%} & \textcolor{dgreen}{-3.88\%} &  & t201 & 185 & 183 & 183 & \textcolor{black}{0.00\%} & \textcolor{dred}{1.08\%} \\
t105 & 433 & 433 & 428 & \textcolor{dred}{1.15\%} & \textcolor{dred}{1.15\%} &  & t202 & 193 & 193 & 191 & \textcolor{dred}{1.04\%} & \textcolor{dred}{1.04\%} \\
t114 & 476 & 467 & 493 & \textcolor{dgreen}{-5.57\%} & \textcolor{dgreen}{-3.57\%} &  & t203 & 179 & 174 & 179 & \textcolor{dgreen}{-2.87\%} & \textcolor{black}{0.00\%} \\
t117 & 462 & 452 & 482 & \textcolor{dgreen}{-6.64\%} & \textcolor{dgreen}{-4.33\%} &  & t204 & 171 & 171 & 176 & \textcolor{dgreen}{-2.92\%} & \textcolor{dgreen}{-2.92\%} \\
t121 & 436 & 424 & 450 & \textcolor{dgreen}{-6.13\%} & \textcolor{dgreen}{-3.21\%} &  & t206 & 201 & 196 & 197 & \textcolor{dgreen}{-0.51\%} & \textcolor{dred}{1.99\%} \\
t122 & 478 & 468 & 461 & \textcolor{dred}{1.50\%} & \textcolor{dred}{3.56\%} &  & t207 & 201 & 174 & 200 & \textcolor{dgreen}{-14.94\%} & \textcolor{dred}{0.50\%} \\
t123 & 409 & 404 & 422 & \textcolor{dgreen}{-4.46\%} & \textcolor{dgreen}{-3.18\%} &  & t208 & 176 & 162 & 176 & \textcolor{dgreen}{-8.64\%} & \textcolor{black}{0.00\%} \\
t124 & 471 & 435 & 468 & \textcolor{dgreen}{-7.59\%} & \textcolor{dred}{0.64\%} &  & t218 & 155 & 155 & 152 & \textcolor{dred}{1.94\%} & \textcolor{dred}{1.94\%} \\
t126 & 415 & 413 & 432 & \textcolor{dgreen}{-4.60\%} & \textcolor{dgreen}{-4.10\%} &  & t223 & 229 & 183 & 228 & \textcolor{dgreen}{-24.59\%} & \textcolor{dred}{0.44\%} \\
t129 & 441 & 432 & 449 & \textcolor{dgreen}{-3.94\%} & \textcolor{dgreen}{-1.81\%} &  & t227 & 159 & 159 & 156 & \textcolor{dred}{1.89\%} & \textcolor{dred}{1.89\%} \\
t131 & 413 & 400 & 413 & \textcolor{dgreen}{-3.25\%} & \textcolor{black}{0.00\%} &  & t229 & 178 & 178 & 182 & \textcolor{dgreen}{-2.25\%} & \textcolor{dgreen}{-2.25\%} \\
t132 & 420 & 420 & 440 & \textcolor{dgreen}{-4.76\%} & \textcolor{dgreen}{-4.76\%} &  & t233 & 212 & 180 & 212 & \textcolor{dgreen}{-17.78\%} & \textcolor{black}{0.00\%} \\
t143 & 417 & 413 & 419 & \textcolor{dgreen}{-1.45\%} & \textcolor{dgreen}{-0.48\%} &  & t236 & 175 & 175 & 177 & \textcolor{dgreen}{-1.14\%} & \textcolor{dgreen}{-1.14\%} \\
t145 & 371 & 357 & 387 & \textcolor{dgreen}{-8.40\%} & \textcolor{dgreen}{-4.31\%} &  & t241 & 172 & 170 & 171 & \textcolor{dgreen}{-0.59\%} & \textcolor{dred}{0.58\%} \\
t148 & 471 & 468 & 475 & \textcolor{dgreen}{-1.50\%} & \textcolor{dgreen}{-0.85\%} &  & t242 & 180 & 180 & 180 & \textcolor{black}{0.00\%} & \textcolor{black}{0.00\%} \\
t150 & 487 & 487 & 485 & \textcolor{dred}{0.41\%} & \textcolor{dred}{0.41\%} &  & t243 & 199 & 170 & 201 & \textcolor{dgreen}{-18.24\%} & \textcolor{dgreen}{-1.01\%} \\
- & - & - & - & - & - &  & t250 & 200 & 200 & 201 & \textcolor{dgreen}{-0.50\%} & \textcolor{dgreen}{-0.50\%} \\
\bottomrule 
\end{tabular}}
\caption{Performance (score, gap to ILS and gap to best-known) on Gavalas instances of a global model trained on all instance-regions of this group and then fine-tuned during 50,000 epochs for each instance-region ("global+ft"). Inference uses beam search with a maximum of 128 beams.}
\label{table:full_results_bench_G}
\end{table}

\subsubsection*{Gavalas [Generated]}
\begin{table}[H]
\renewcommand{\arraystretch}{1.5} 
\centering
\resizebox{.5\width}{!}{\rowcolors{2}{black!5}{white}
\begin{tabular}{ccccccccc}
\toprule 
\textbf{instance [generated]} & \textbf{ILS} & \textbf{global+ft} & \textbf{global+ft gap to ILS} &  & \textbf{instance [generated]} & \textbf{ILS} & \textbf{global+ft} & \textbf{global+ft gap to ILS} \\
\midrule 
t101 & 320.44 & 339.11 & \textcolor{dgreen}{-5.83\%} &  & t201 & 235.88 & 265.70 & \textcolor{dgreen}{-12.65\%} \\
t105 & 363.44 & 392.48 & \textcolor{dgreen}{-7.99\%} &  & t202 & 252.59 & 302.25 & \textcolor{dgreen}{-19.66\%} \\
t114 & 328.36 & 358.50 & \textcolor{dgreen}{-9.18\%} &  & t203 & 250.39 & 273.83 & \textcolor{dgreen}{-9.36\%} \\
t117 & 401.27 & 455.58 & \textcolor{dgreen}{-13.54\%} &  & t204 & 247.53 & 278.66 & \textcolor{dgreen}{-12.57\%} \\
t121 & 346.19 & 390.00 & \textcolor{dgreen}{-12.66\%} &  & t206 & 327.95 & 365.36 & \textcolor{dgreen}{-11.41\%} \\
t122 & 373.58 & 400.97 & \textcolor{dgreen}{-7.33\%} &  & t207 & 248.69 & 278.41 & \textcolor{dgreen}{-11.95\%} \\
t123 & 301.97 & 338.45 & \textcolor{dgreen}{-12.08\%} &  & t208 & 275.28 & 308.11 & \textcolor{dgreen}{-11.93\%} \\
t124 & 332.53 & 357.98 & \textcolor{dgreen}{-7.65\%} &  & t218 & 230.44 & 251.95 & \textcolor{dgreen}{-9.34\%} \\
t126 & 356.33 & 380.17 & \textcolor{dgreen}{-6.69\%} &  & t223 & 263.16 & 294.91 & \textcolor{dgreen}{-12.07\%} \\
t129 & 326.97 & 360.83 & \textcolor{dgreen}{-10.36\%} &  & t227 & 260.58 & 287.25 & \textcolor{dgreen}{-10.24\%} \\
t131 & 327.12 & 351.38 & \textcolor{dgreen}{-7.41\%} &  & t229 & 262.86 & 295.47 & \textcolor{dgreen}{-12.41\%} \\
t132 & 351.36 & 394.22 & \textcolor{dgreen}{-12.20\%} &  & t233 & 251.92 & 289.14 & \textcolor{dgreen}{-14.77\%} \\
t143 & 368.66 & 408.88 & \textcolor{dgreen}{-10.91\%} &  & t236 & 241.67 & 268.73 & \textcolor{dgreen}{-11.20\%} \\
t145 & 320.66 & 349.05 & \textcolor{dgreen}{-8.85\%} &  & t241 & 268.17 & 296.42 & \textcolor{dgreen}{-10.53\%} \\
t148 & 342.66 & 372.66 & \textcolor{dgreen}{-8.76\%} &  & t242 & 249.20 & 273.86 & \textcolor{dgreen}{-9.89\%} \\
t150 & 394.69 & 432.02 & \textcolor{dgreen}{-9.46\%} &  & t243 & 236.42 & 269.89 & \textcolor{dgreen}{-14.16\%} \\
- & - & - & - &  & t250 & 264.53 & 315.47 & \textcolor{dgreen}{-19.26\%} \\
\bottomrule 
\end{tabular}}
\caption{Average performance (score and gap to ILS) on Gavalas generated instances of a global model trained on all instance-regions of this group and then fine-tuned during 50,000 epochs for each instance-region ("global+ft"). Inference uses beam search with a maximum of 128 beams. The values are averages across the n=64 generated instances.}
\label{table:full_results_gener_G}
\end{table}

\section{Benchmark Instances Data}
\label{app:instance_stats}
In this section we present visualizations that should give information about the characteristics of each benchmark instance as well as the variability withing and across each instance group.
\subsection{Spacial Distribution}
\subsubsection*{Solomon}

\begin{figure}[H]
\includegraphics[width=0.95\textwidth]{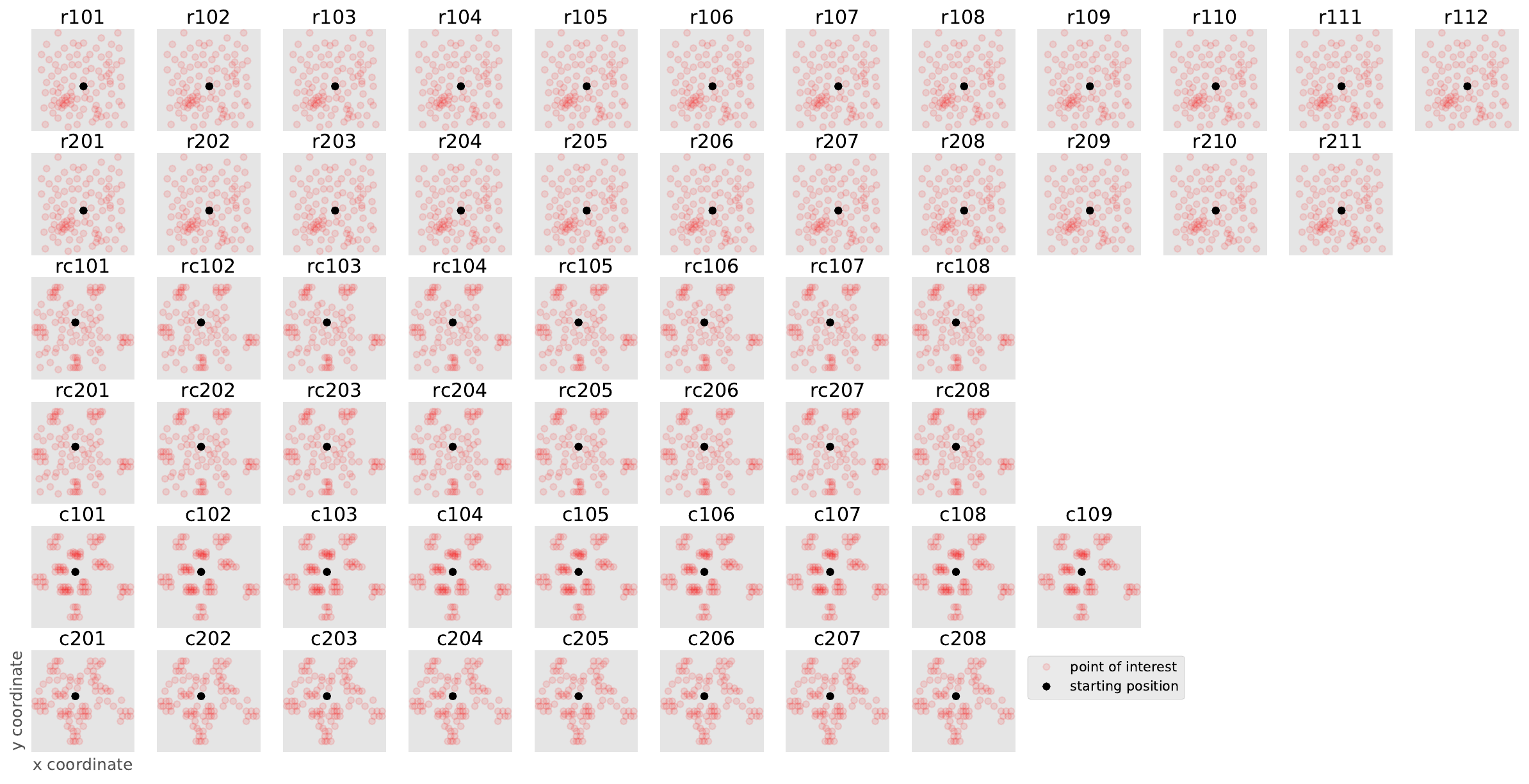}
\centering
\caption{Spacial distribution of the points of interest (red), and starting and ending location (black) for the benchmark instances in the Solomon group.}
\label{fig:stats_location_solomon}
\end{figure}

\subsubsection*{Cordeau}

\begin{figure}[H]
\includegraphics[width=0.95\textwidth]{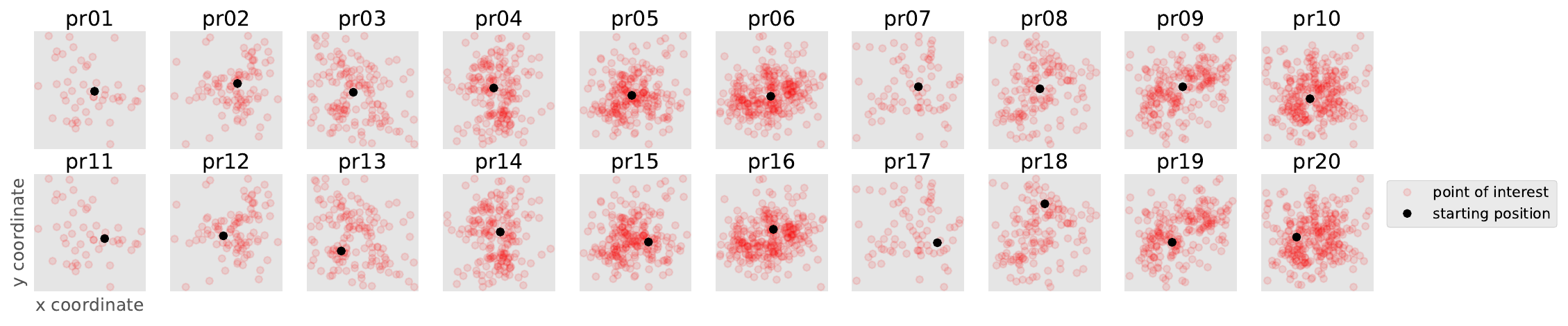}
\centering
\caption{Spacial distribution of the points of interest (red), and starting and ending location (black) for the benchmark instances in the Cordeau group.}
\label{fig:stats_location_cordeau}
\end{figure}

\subsubsection*{Gavalas}

\begin{figure}[H]
\includegraphics[width=0.95\textwidth]{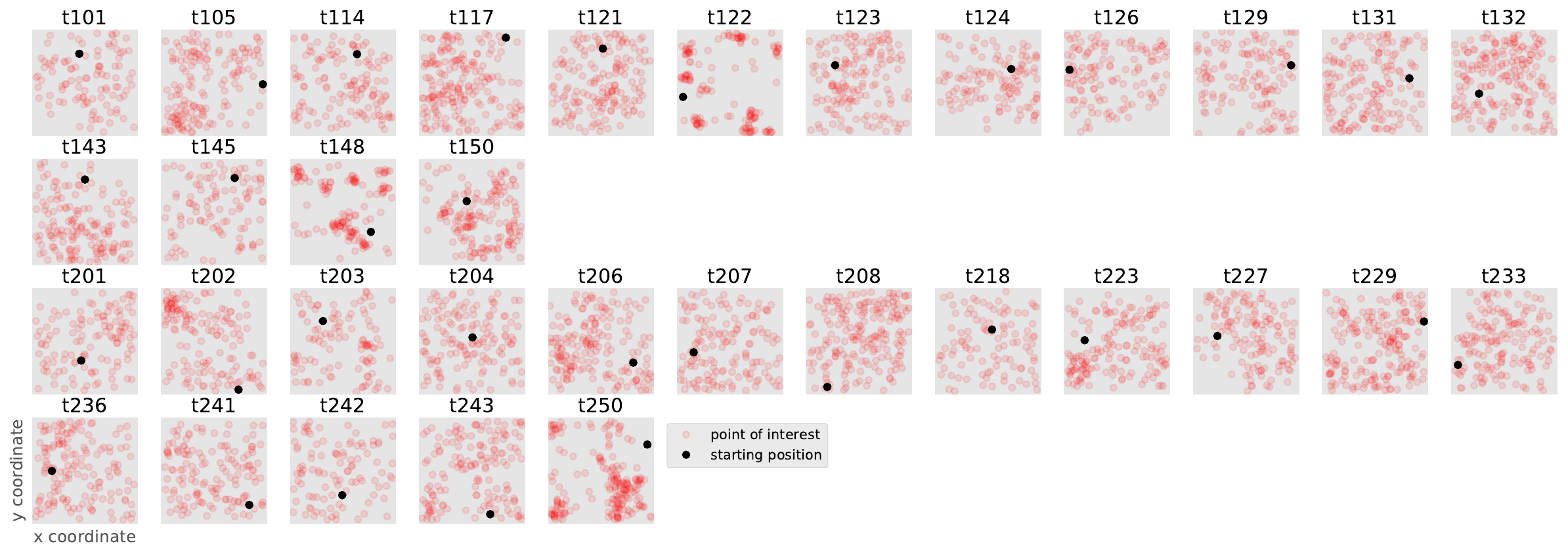}
\centering
\caption{Spacial distribution of the points of interest (red), and starting and ending location (black) for the benchmark instances in the Gavalas group.}
\label{fig:stats_location_gavalas}
\end{figure}

\subsection{Correlation between Scores and Duration of Visit}

\subsubsection*{Solomon}

\begin{figure}[H]
\includegraphics[width=0.95\textwidth]{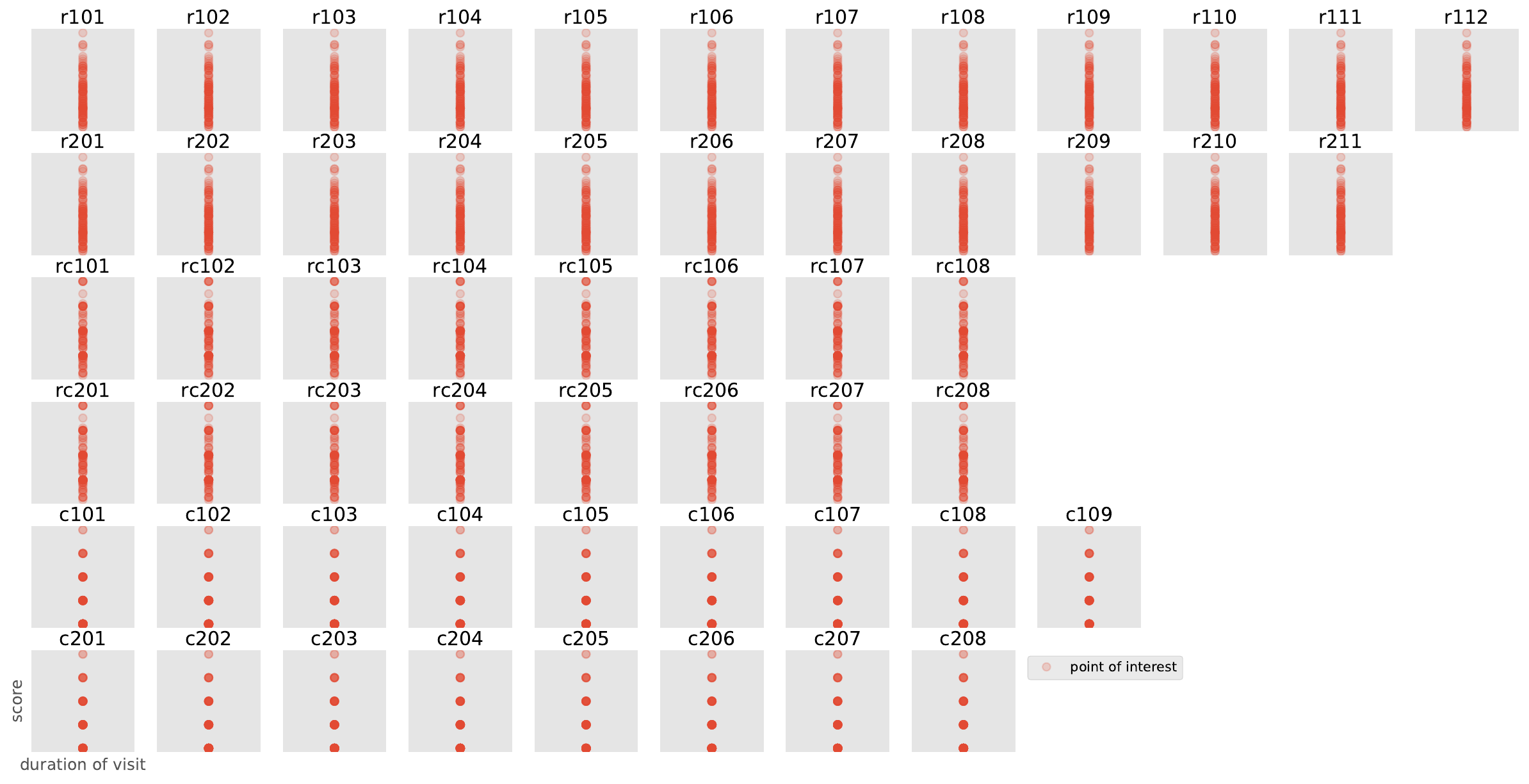}
\centering
\caption{Score of each point of interest as a function of the time it takes to visit that point of interest for each benchmark instance in the Solomon group.}
\label{fig:stats_visit_dur_vs_score_solomon}
\end{figure}

\subsubsection*{Cordeau}

\begin{figure}[H]
\includegraphics[width=0.95\textwidth]{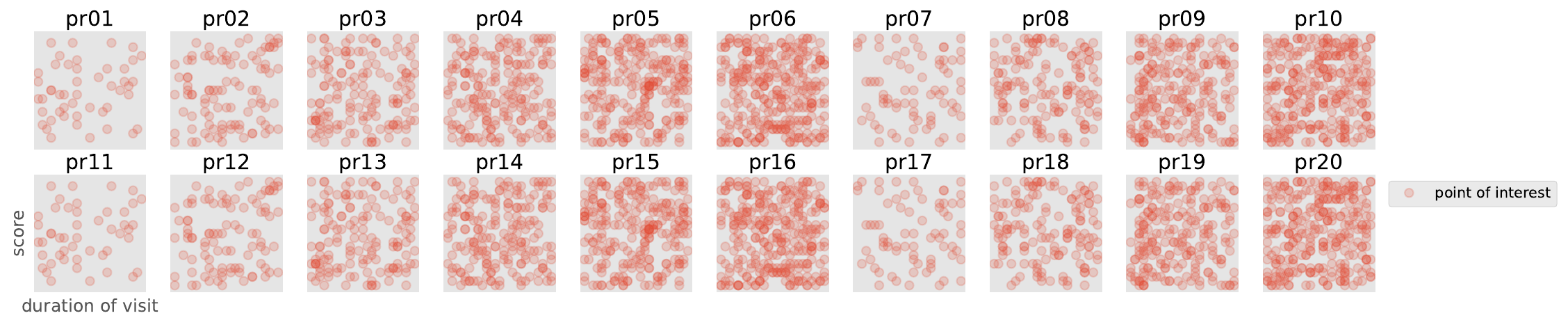}
\centering
\caption{Score of each point of interest as a function of the time it takes to visit that point of interest for each benchmark instance in the Cordeau group.}
\label{fig:stats_visit_dur_vs_score_cordeau}
\vspace*{1in}
\end{figure}

\subsubsection*{Gavalas}

\begin{figure}[H]
\includegraphics[width=0.95\textwidth]{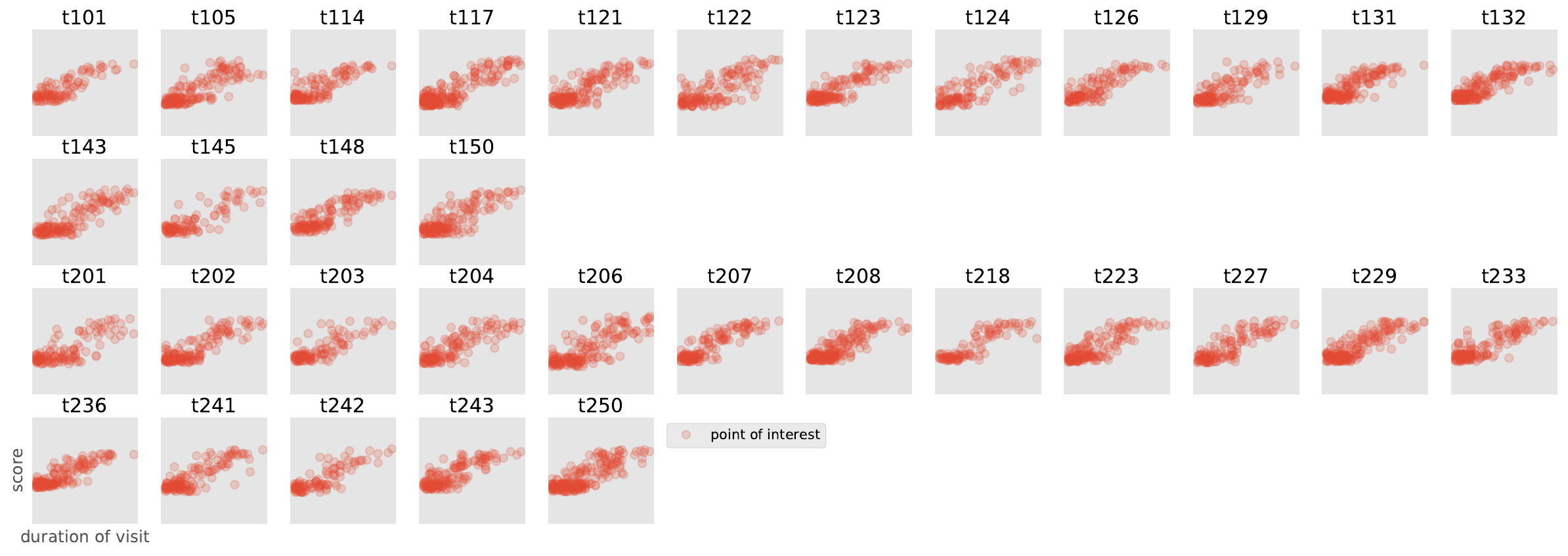}
\centering
\caption{Score of each point of interest as a function of the time it takes to visit that point of interest for each benchmark instance in the Gavalas group.}
\label{fig:stats_visit_dur_vs_score_gavalas}
\vspace*{1in}
\end{figure}

\subsection{Schedules}

\subsubsection*{Solomon}

\begin{figure}[H]
\includegraphics[width=0.75\textwidth]{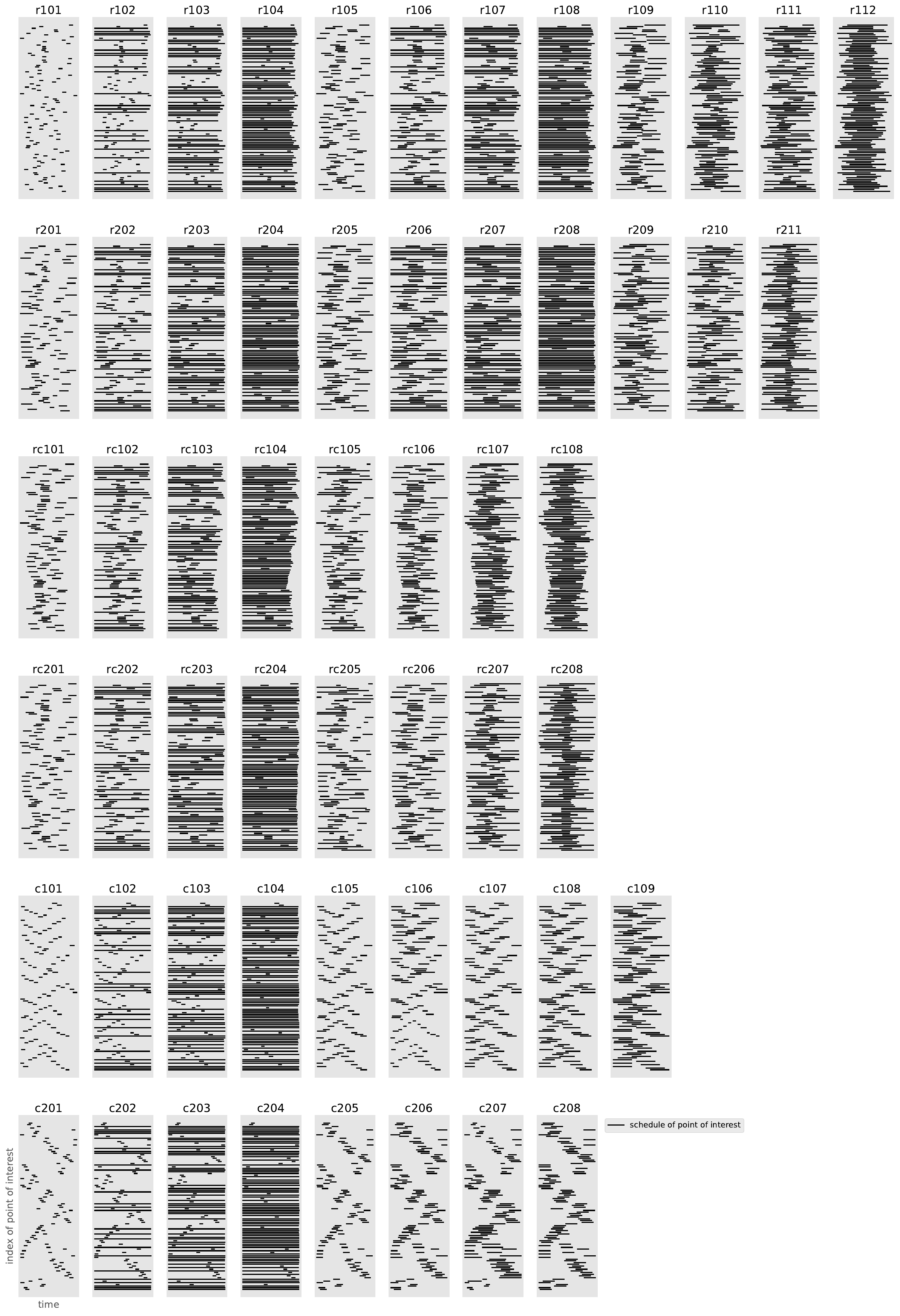}
\centering
\caption{Schedule (opening time to closing time) of each point of interest for all benchmark instances of the Solomon group.}
\label{fig:stats_schedules_solomon}
\end{figure}

\subsubsection*{Cordeau}

\begin{figure}[H]
\includegraphics[width=0.75\textwidth]{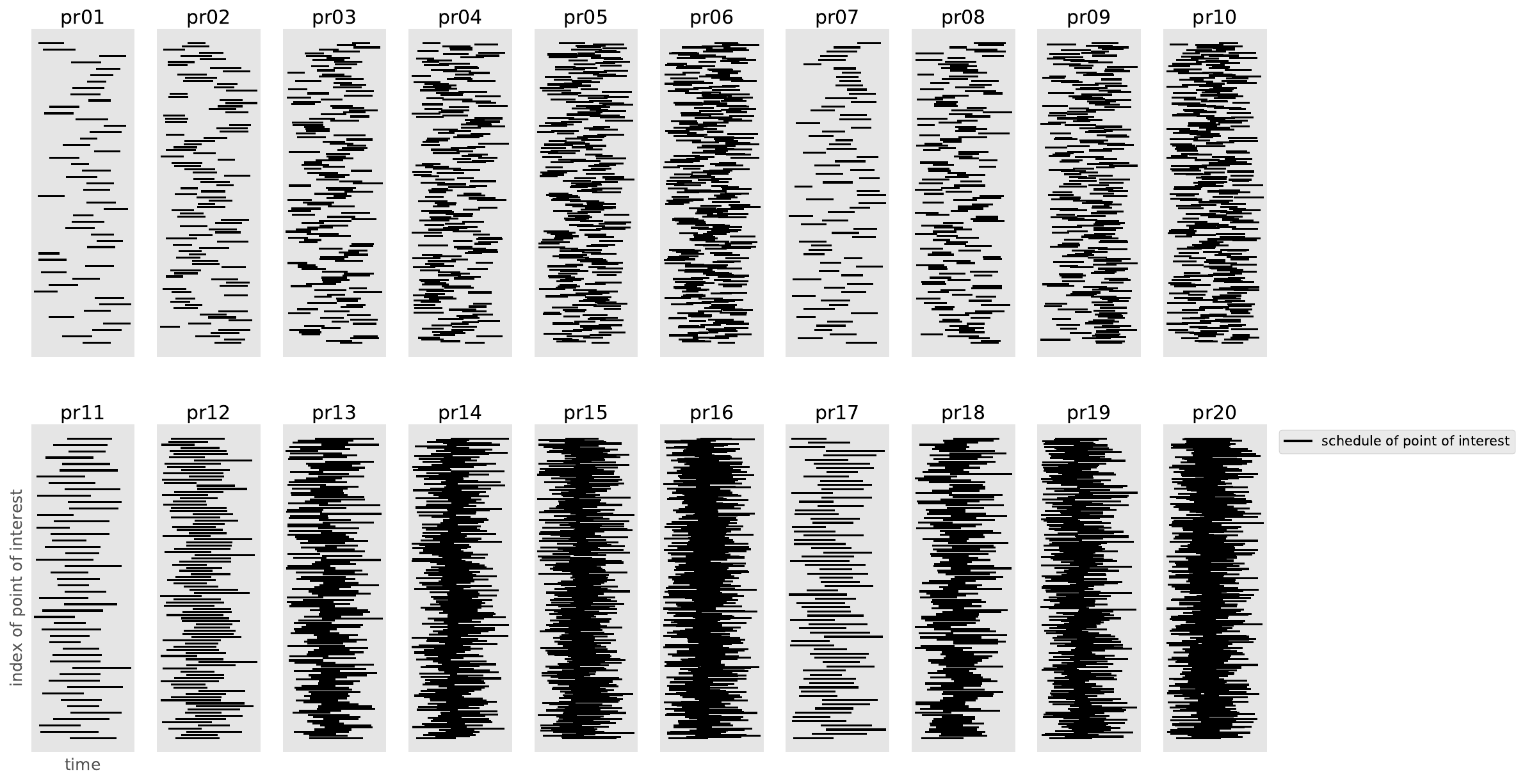}
\centering
\caption{Schedule (opening time to closing time) of each point of interest for all benchmark instances of the Cordeau group.}
\label{fig:stats_schedules_cordeau}
\end{figure}

\subsubsection*{Gavalas}

\begin{figure}[H]
\includegraphics[width=0.75\textwidth]{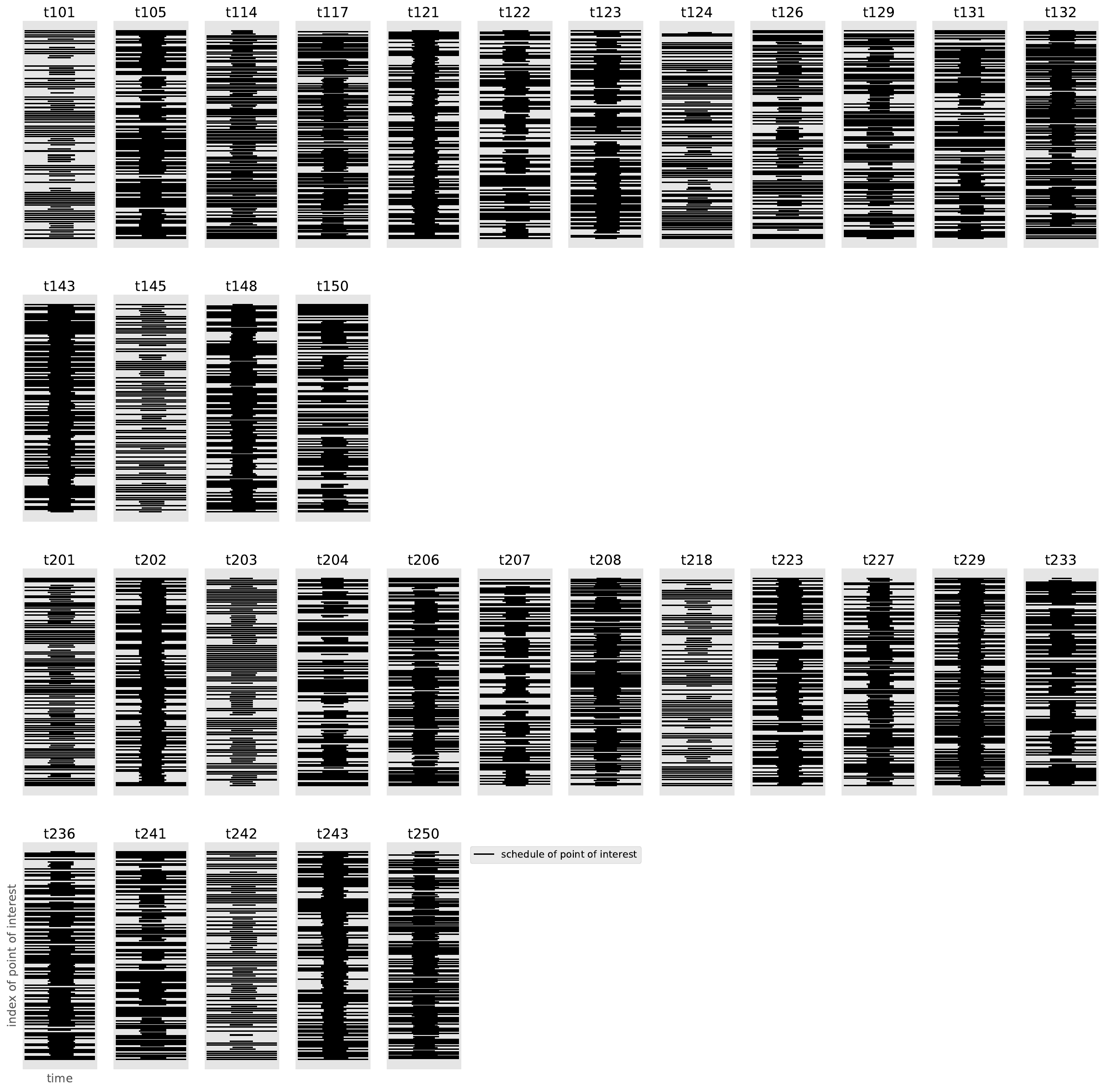}
\centering
\caption{Schedule (opening time to closing time) of each point of interest for all benchmark instances of the Gavalas group.}
\label{fig:stats_schedules_gavalas}
\end{figure}

\end{document}